\documentclass{article}

\usepackage{microtype}
\usepackage{graphicx}
\usepackage{subfigure}
\usepackage{booktabs} 

\usepackage{hyperref}

\usepackage[accepted]{icml2024}

\usepackage{amsmath}
\usepackage{amssymb}
\usepackage{mathtools}
\usepackage{amsthm}

\usepackage[capitalize,noabbrev]{cleveref}

\theoremstyle{plain}

\theoremstyle{definition}

\theoremstyle{remark}

\usepackage[textsize=tiny]{todonotes}
\usepackage{hyperref}
\usepackage{colortbl}
\usepackage{xspace}

\newcommand{\xhdr}[1]{\vspace{1.7mm}\noindent{{\bf #1.}}}

\newcommand{\method}{{CROW}\xspace}

\icmltitlerunning{Cross-domain Open-world Discovery}

\begin{document}

\twocolumn[
\icmltitle{Cross-domain Open-world Discovery}




\begin{icmlauthorlist}
\icmlauthor{Shuo Wen}{to}
\icmlauthor{Maria Brbić}{to}
\end{icmlauthorlist}

\icmlaffiliation{to}{EPFL, Switzerland}

\icmlcorrespondingauthor{Maria Brbić}{mbrbic@epfl.ch}

\icmlkeywords{Machine Learning, ICML}

\vskip 0.3in
]



\printAffiliationsAndNotice{}  

\begin{abstract}
In many real-world applications, test data may commonly exhibit categorical shifts, characterized by the emergence of novel classes, as well as distribution shifts arising from feature distributions different from the ones the model was trained on. However, existing methods either discover novel classes in the open-world setting or assume domain shifts without the ability to discover novel classes. In this work, we consider a \textit{cross-domain open-world discovery} setting, where the goal is to assign samples to seen classes and discover unseen classes under a domain shift. To address this challenging problem, we present \method, a prototype-based approach that introduces a cluster-then-match strategy enabled by a well-structured representation space of foundation models. In this way, \method discovers novel classes by robustly matching clusters with previously seen classes, followed by fine-tuning the representation space using an objective designed for cross-domain open-world discovery. Extensive experimental results on image classification benchmark datasets demonstrate that \method outperforms alternative baselines, achieving an $8\%$ average performance improvement across $75$ experimental settings.
\end{abstract}

\section{Introduction}
\label{sec:intro}
\begin{figure}[t]
    \centering
    \includegraphics[width=\linewidth]{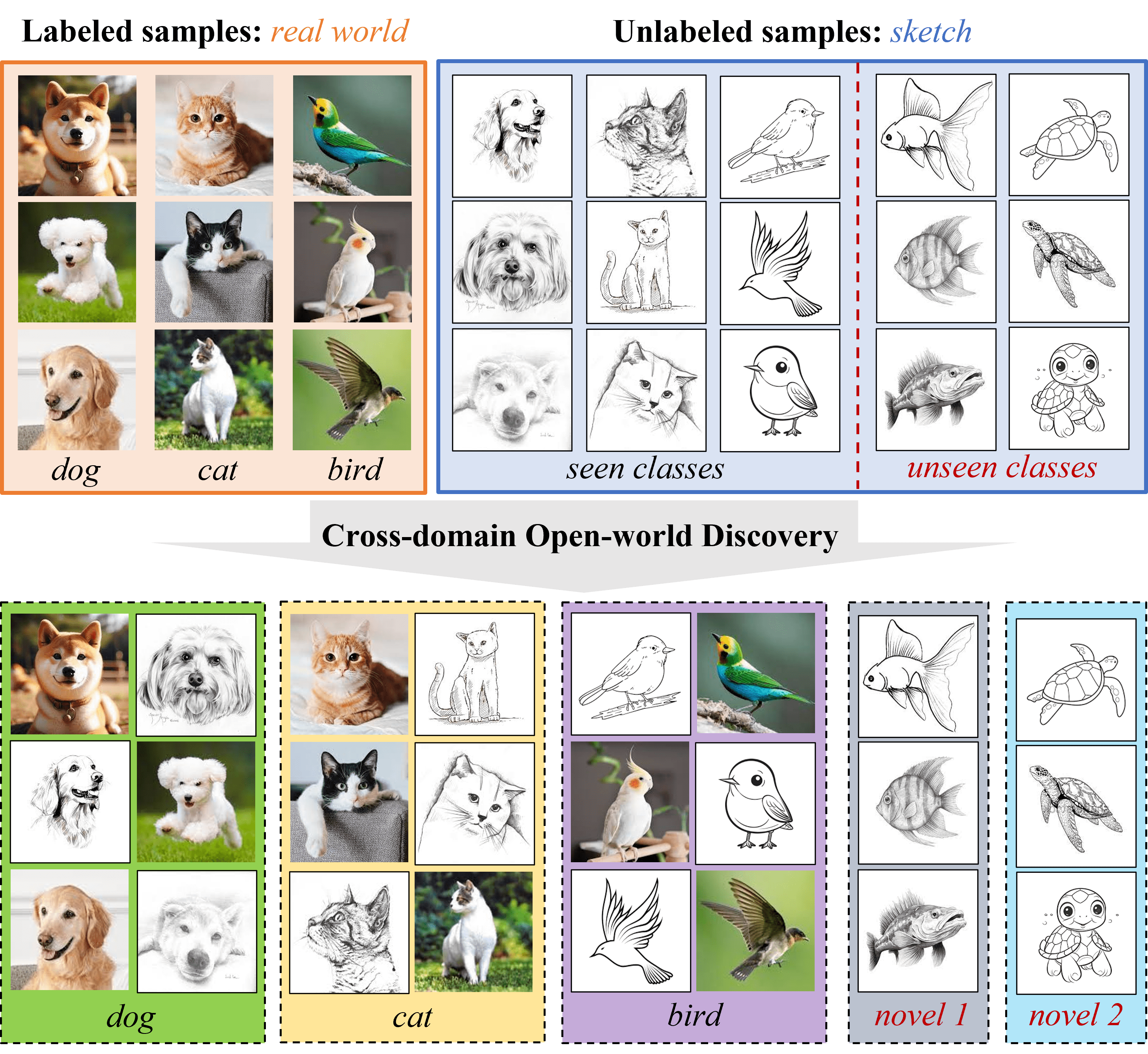}
    \vspace{-6mm}
    \caption{\textbf{Illustration of the cross-domain open-world discovery setting.} In the cross-domain open-world discovery setting, the goal is to assign samples to previously seen classes and discover new classes under a domain shift. In the example, novel classes like \textit{`fish'} and \textit{`turtle'}, exist in unlabeled data. Additionally, the labeled samples are from the real-world domain, while the unlabeled samples are sketches. In this setting, the goal is to assign each unlabeled sample to either a seen category (\textit{`dog'}, \textit{`cat'}, \textit{`bird'}) or to a novel category that is discovered (\textit{`novel 1'}, \textit{`novel 2'}).} 
    \label{fig:intro}
\end{figure}

The rise of deep learning has brought significant advancements, empowering machine learning systems with exceptional performance in tasks requiring extensive labeled data~\cite{lecun2015deep, schmidhuber2015deep, silver2016mastering}. However, many models are developed within a closed-world paradigm, assuming that training and test data originate from a predetermined set of classes within the same domain. 
This assumption is overly restrictive in many real-world scenarios. 
For example, a model trained to categorize diseases in medical images from one hospital may experience domain shifts when applied to images from different hospitals. Moreover, during model deployment, novel and rare diseases may emerge that the model has never seen during training.
In the open-world scenario, the model should have the capability to generalize beyond predefined classes and domains, a departure from the closed-world scenario often presumed in traditional approaches.

Open-world learning~\cite{open1} extends beyond closed-world paradigms by enabling models to recognize unseen classes and scenarios, addressing the dynamic challenges of real-world environments.
In this context, open-world semi-supervised learning (OW-SSL)~\cite{orca} defines a setting in which the objective is to annotate seen classes and discover unseen classes. 
However, OW-SSL assumes that the labeled and unlabeled data belong to the same domain, which is often not the case. 
On the other hand, Universal Domain Adaptation (UniDA)~\cite{unida} tackles the problem of domain and categorical shifts between labeled and unlabeled data. However, the primary objective of UniDA is to assign samples to seen classes and reject unseen samples as outliers, rather than discover novel unseen classes. 

In this work, we address this gap by considering a Cross-Domain Open-World Discovery (CD-OWD) setting.
In this setting, the objective is to assign samples to pre-existing (seen) classes while simultaneously being able to discover new (unseen) classes under a domain shift (Figure~\ref{fig:intro}).
This setting operates within a transductive learning framework, where we have access to both a labeled dataset (source set) and an unlabeled dataset (target set) during training. In contrast to OW-SSL, this setting considers not only categorical shifts but also domain shifts.
In contrast to UniDA, the goal here is to discover novel classes instead of rejecting all of them as unknowns. Thus, CD-OWD needs to overcome the challenges of both open-world semi-supervised learning and universal domain adaptation. This setting has been previously considered in ~\cite{rebuttal1}. However, their evaluation approach is not suitable for the proposed setting hindering the ability to effectively solve the proposed task. 

A straightforward approach to tackle this challenge is to first apply one of the UniDA methods~\cite{dance,ovanet,uniot,glc} to annotate the seen samples and identify the unseen samples. After that, the detected unseen samples can be clustered to discover novel classes. We call this approach \textit{match-then-cluster}.
In practice, this approach encounters two problems. First, UniDA methods rely on a sensitive threshold to separate seen and unseen samples. Finding the optimal threshold using validation sets is not feasible because the domain gap between labeled and unlabeled samples prevents the creation of validation sets that accurately reflect the target domain. Second, when UniDA methods fail to perfectly separate seen and unseen samples, the seen samples misclassified as unseen introduce noise to the unseen samples, thereby reducing the quality of the clustering process.

To overcome these challenges, we propose \method (\textbf{C}ross-domain \textbf{R}obust \textbf{O}pen-\textbf{W}orld-discovery), a method that employs a \textit{cluster-then-match} approach, leveraging the capabilities of foundation models. The key idea in \method is to utilize the well-structured latent space of foundation models \cite{clip, dinov2,swag} to first cluster the data and then use a robust prototype-based matching strategy. This matching strategy enables \method \space to associate multiple target prototypes with seen classes, thereby alleviating the issues of over-clustering and under-clustering.
After matching prototypes, \method combines cross-entropy loss applied to source samples with entropy maximization loss applied to target samples to further improve the representation space.

We evaluate \method across $75$ different categorical-shift and domain-shift scenarios created from four benchmark domain adaptation datasets for image classification. 
The results demonstrate that our approach outperforms open-world semi-supervised learning and universal domain adaptation baselines by a large margin. Specifically, CROW outperforms the strongest baseline GLC by an average of $8\%$ on the H-score. Moreover, CROW is robust to different hyperparameters, an unknown number of target classes, and different seen/unseen splits.

\section{Related work}
\label{sec:related_work}
The cross-domain open-world discovery setting is closely related to open-world semi-supervised learning and universal domain adaptation. It is a harder setting compared to these two settings as it requires overcoming the challenges of both settings –– we need to discover novel classes under a domain shift. \method builds upon the power of foundation models, allowing us to adopt the \textit{cluster-then-match} strategy proposed in this work.

\xhdr{Open-world learning} 
Open-world learning~\cite{open1,open2,open3} entails annotating unlabeled data in the face of categorical shift, where new classes may arise in the unlabeled data. 
Open Set Label Shift (OSLS)~\cite{pulse} is a setting that detects the samples from the seen classes and annotates them. However, it focuses on seen classes and does not separate different unseen classes.
Novel Class Discovery (NCD)~\cite{ncd} aims to discover unseen classes. However, NCD assumes that all the unlabeled samples are from novel classes, so it does not need to detect common classes.
Open-world semi-supervised learning (OW-SSL)~\cite{orca} combines the settings of OSLS and NCD. It aims to annotate seen classes and discover unseen classes under the assumption that the unlabeled samples are from both seen and novel classes. 
However, OW-SSL assumes that labeled and unlabeled data belong to the same domain, which is not always true. 
In this work, we consider the \textit{cross-domain open-world discovery} setting which accounts for domain shift.

\xhdr{Unsupervised domain adaptation}\label{sec:uda}
Unsupervised domain adaptation (UDA)~\cite{UDA} aims to annotate unlabeled data under domain shift between labeled and unlabeled data. However, it assumes that labeled and unlabeled data originate from the same classes.
Open-Set Domain Adaptation (OSDA)~\cite{osda1} and
Universal Domain Adaptation (UniDA)~\cite{unida} extend the setting of UDA by considering unseen classes in the unlabeled data.
They aim to annotate seen classes and detect unseen samples. 
Prior works \cite{osda2, unida2, ovanet, unida_cam, rebuttal2} achieved significant success within both the OSDA and UniDA setting.
However, these works reject unseen samples without exploring the internal structure of the unseen part.
Recent works \cite{dance, dcc, osda3, uniot,  unida_iclr24} have started paying attention to the internal structure of the target domain, especially for the unknown samples. However, most of them explore internal structures to better detect unseen samples but not to separate them according to new classes. 
UniDA methods are applicable to the cross-domain discovery setting by clustering the samples detected as novel.
\citet{rebuttal1} consider the cross-domain discovery setting. However, their evaluation strategy is not suitable since it does not directly evaluate the ability to discover novel classes. As a result, the evaluation does not accurately reflect performance in the context of the cross-domain discovery setting, leaving its effectiveness unclear.

We compare different problem settings in Table~\ref{table:problem setting}.

\vspace{-5mm}
\begin{table}[h]
\caption{Comparison of different problem settings. UDA stands for Universal Domain Adaptation; OSLS for Open Set Label Shift; NCD for Novel Class Discovery; UniDA for Universal Domain Adaptation; OW-SSL for Open-World Semi-Supervised Learning; and CD-OWD for Cross-Domain Open-World Discovery.}
\vspace{-10pt}
\label{table:problem setting}
\begin{center}
\resizebox{\linewidth}{!}{
\begin{small}
\begin{sc}
\begin{tabular}{l|ccc}
\toprule
Setting  &  Domain Shift & Seen Detection & Novel Discovery\\
\midrule
UDA  & $\surd$ & - & -\\
OSLS & - & $\surd$ & -\\
NCD  & - & - & $\surd$\\
UniDA  & $\surd$ & $\surd$ & -\\
OW-SSL  & - & $\surd$ & $\surd$\\
\textbf{CD-OWD}  & $\surd$ & $\surd$ & $\surd$\\
\bottomrule
\end{tabular}
\end{sc}
\end{small}
}
\end{center}
\vskip -0.1in
\end{table}

\xhdr{Transfer learning and foundation models} 
Existing open-world and domain adaptation methods generally use the standard pretraining and fine-tuning paradigm of transfer learning~\cite{transfer3,transfer2,bit}. 
The pretrained feature extractors provide a well-structured latent space, allowing faster training and better generalization. Previous works on OW-SSL and UniDA~\cite{dance,ovanet,uniot,glc,orca} directly use the ImageNet~\cite{imagenet} supervised pretrained or SimCLR~\cite{simclr} self-supervised pretrained ResNet50~\cite{resnet} as their backbone.
However, recently developed foundation models ~\cite{clip,swag,dinov2} provide a better structured initial latent space, eliminating the necessity for self-supervised training on target data to achieve reliable initialization.
Previous works~\cite{unida_foundation,da_foundation,da_foundation2} show that foundation models help alleviate domain shifts in their representation space.
In this work, we build our method upon the power of the well-structured representation space of foundation models. This allows us to adopt a \textit{cluster-then-match} strategy in contrast to the \textit{match-then-cluster} strategy extended from existing universal domain adaptation methods.

\xhdr{Cluster-then-match approach} To solve the universal domain adaptation problem, DCC~\cite{dcc} first clusters the unlabeled target samples and then matches each target cluster to one seen class for recognizing target seen classes. This approach corresponds to the strategy we call \textit{cluster-then-match} in this work. 
However, a limitation of this specific instantiation of the \textit{cluster-then-match} strategy is that DCC requires one-to-one matching between seen classes and target clusters. This requirement cannot be satisfied in the condition of under-clustering (\textit{i.e.}, assigning multiple seen classes to the same cluster) and over-clustering (\textit{i.e.}, splitting a single seen class into multiple clusters), as the relationship between seen classes and target clusters is no longer one-to-one. 
Instead, we adopt robust matching, which mitigates this problem by releasing the constraint of one-to-one matching, allowing multiple seen classes to be matched to the same cluster and a single seen class to be matched to multiple clusters.

\section{Method}
\label{sec:method}
\begin{figure*}[ht]
\begin{center}
\centerline{\includegraphics[width=1\linewidth]{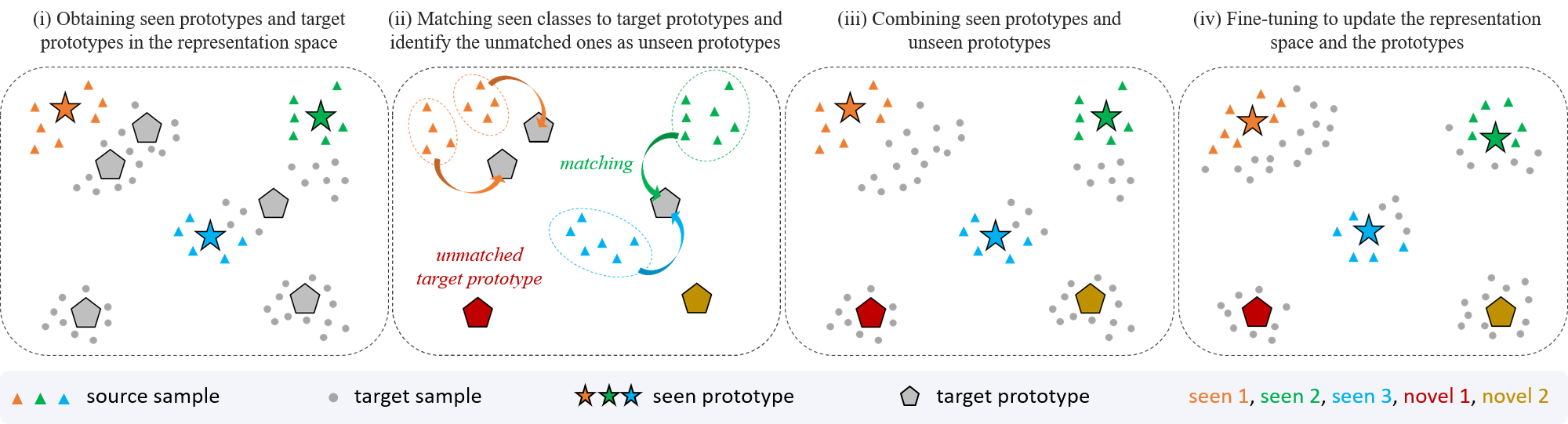}}
\vskip -0.1in
\caption{\textbf{Conceptual overview of \method.} \textit{(i)} \method extracts features from a foundation model for both source and target samples. Seen prototypes are then obtained using labeled source samples, while target prototypes are obtained by clustering target samples. \textit{(ii)} \method matches seen classes to target prototypes using the source samples. Unmatched target prototypes are identified as unseen prototypes. \textit{(iii)} \method combines seen prototypes and unseen prototypes. \textit{(iv)} Finally, \method fine-tunes the foundation model to update the representation space and the prototypes.}
\label{fig:concept}
\end{center}
\vskip -0.3in
\end{figure*}

\subsection{Cross-domain open-world discovery setting}
\label{sec:setting}

In the cross-domain open-world discovery, we assume a transductive learning setting, where a labeled dataset (\textit{i.e.}, source set) $D_s=\{(x_i,y_i)\}^n_{i=1}$ and an unlabeled dataset (\textit{i.e.}, target set) $D_t=\{x_i\}^m_{i=1}$ are given during training. We denote the set of classes in the source set as $C_s$ and the set of classes in the target set as $C_t$. We consider both the categorical shift and the domain shift. 
Under the \textbf{categorical shift}, we assume $C_s\cap C_t \neq \emptyset$ and $C_s\neq C_t$. We consider $C_s$ as a set of seen classes and $C_t\setminus C_s$ as a set of novel classes. Additionally, under the \textbf{domain shift}, we consider $P(x)$ as the feature distribution of data $x$. We assume that $P(x^s)\neq P(x^t)$, where $x^s\in D_s$ and $x^t\in D_t$.

The objective is to assign each $x_i\in D_t$ a label $y_i$. The $y_i$ is either from a seen class in $C_s$ or from a novel class that is discovered.

\subsection{Overview of \method}
\label{sec:overview}

To overcome the challenges of cross-domain open-world discovery, the novel classes need to be well separated in the representation space. Early incorporation of labels from seen classes in the training process can lead to a bias towards seen classes, hindering the ability to differentiate between samples of novel classes. The key idea in \method is to adopt the \textit{cluster-then-match} strategy, enabled by the well-structured representation space of foundation models. In particular, \method first clusters the target samples in the representation space of a foundation model, followed by a robust matching that associates seen classes with the target clusters. Finally, CROW is fine-tuned using an objective specially tailored for the open-world discovery setting. 

Thus, \method adopts a three-step procedure approach, including: \textit{(i)} clustering, \textit{(ii)} matching, and \textit{(iii)} fine-tuning. 

\subsection{Clustering step}
\label{sec:clustering}

To obtain clusters of target samples, \method leverages 
a robust representation space of a foundation model~\cite{clip, swag, dinov2, gadetsky2024pursuit}. The goal of the clustering step in \method is to obtain the prototypes of target sample clusters (referred to as target prototypes) and the prototypes of the seen classes (referred to as seen prototypes).

The foundation model is used as a feature extractor $f_{\theta}$. Let $\mathcal{X}$ be the input space; the feature extractor $f_{\theta}: \mathcal{X}\to\mathbb{R}^d$ maps the input space  $\mathcal{X}$ to a $d$-dimensional representation space. Specifically, given input $x\in\mathcal{X}$, $f_{\theta}$ extracts the feature $z\in\mathbb{R}^d$ by $z = f_{\theta}(x)$.
Note that we add an $L_2$-normalized layer at the end, so the feature $z$ is $L_2$-normalized.

To get the seen prototypes $W_{seen} = [p^s_1, p^s_2, ..., p^s_{|C_s|}]$, where $[\cdot]$ denotes concatenation, we consider $W_{seen}$ as a $L_2$-normalized linear classifier and train it on top of the representation space of a foundation model $f_{\theta}$. In particular, we optimize cross-entropy loss on source samples to obtain seen prototypes $W_{seen}$. Specifically, for each source sample $x^s$, we first extract the feature $z^s$ by $z^s = f_{\theta}(x^s)$. Then, we obtain the predictions using $p(y|x^s) = \sigma(W_{seen}^T\cdot z^s)$, where $\sigma$ is the softmax activation function. Finally, we optimize $W_{seen}$ by applying cross-entropy loss on $p(y|x^s)$. Note that we freeze the feature extractor $f_\theta$ during this process.

To obtain the target prototypes $W_t = [p^t_1, p^t_2, ..., p^t_{|C_t|}]$, we first extract the features of all target samples using $z^t = f_\theta(x^t)$. Then, we apply a K-means clustering with $k=|C_t|$ to get the target prototypes $W_t = [p^t_1, p^t_2, ..., p^t_{|C_t|}]$. Here, we assume the number of target classes $|C_t|$ is given as a prior.

The clustering step of \method, which results in seen prototypes $W_{seen}$ and target prototypes $W_t$ is illustrated in Figure~\ref{fig:concept} \textit{(i)}.
After obtaining the seen prototypes $W_{seen}$, the goal is to identify the unseen prototypes $W_{unseen}$ from target prototypes $W_t$ in the matching step. 

\subsection{Matching step}
\label{sec:matching}

In the matching step of \method, the goal is to identify target prototypes that belong to unseen classes. This is achieved by matching seen classes to target prototypes and designating unmatched target prototypes as the prototypes of unseen classes. To accomplish this, \method employs on a robust matching procedure that allows multiple seen classes to match a target prototype and multiple target prototypes to match a seen class.

To match seen classes to target prototypes, we first compute a co-occurrence matrix $\Gamma\in\mathbb{R}^{|C_t|\times|C_s|}$ between target prototypes and seen classes. This co-occurrence matrix represents the number of source samples from a given seen class assigned to a target prototype. We assign a source sample to a target prototype if that prototype is its nearest prototype in the representation space. 
In essence, the co-occurrence matrix $\Gamma$ quantifies the proximity of seen classes to the target prototypes.

After computing the co-occurrence matrix $\Gamma$, we apply a column-wise softmax to $\Gamma$ and obtain the distribution matrix $D$ as follows:
\begin{equation}
    D_{i,j} = \dfrac{e^{\Gamma_{i,j}}}{\sum_{k = 1}^{|C_t|} e^{\Gamma_{k,j}}}.
\end{equation}

Each column of $D$ represents the distribution of the source samples of a seen class to the target prototypes. Finally, we obtain the matching matrix $M$ by applying a threshold $\tau$ to $D$:

\begin{equation}
\label{equ:match}
M_{i,j} = 
    \left\{ 
  \begin{array}{ c l }
    1 & \quad D_{i,j} \geq \tau \\
    0 &  \quad D_{i,j} < \tau
  \end{array}
\right.
.
\end{equation}

Here, $M_{i,j} = 1$ means that the seen class $C_j$ is matched to target prototype $p^t_i$. 
After matching seen classes to target prototypes, we can easily identify the target prototypes that have not been matched to any seen class as the prototypes of unseen classes. This step is illustrated in Figure~\ref{fig:concept} \textit{(ii)}.

The matching step gives us the unseen prototypes $W_{unseen}$, and we have already obtained the seen prototypes $W_{seen}$ from the clustering step. We then combine them to initialize a linear classifier $W = [W_{seen}, W_{unseen}]$ on top of the feature extractor $f_{\theta}$. This step is illustrated in Figure~\ref{fig:concept} \textit{(iii)}.

Note that the number of identified unseen prototypes may not necessarily be equal to $(|C_t|-|C_s|)$. This disparity arises because the number of matched target prototypes can differ from the number of seen classes $|C_s|$ due to under-clustering and over-clustering issues.

\begin{figure}[h]
    \centering
    \includegraphics[width=\linewidth]{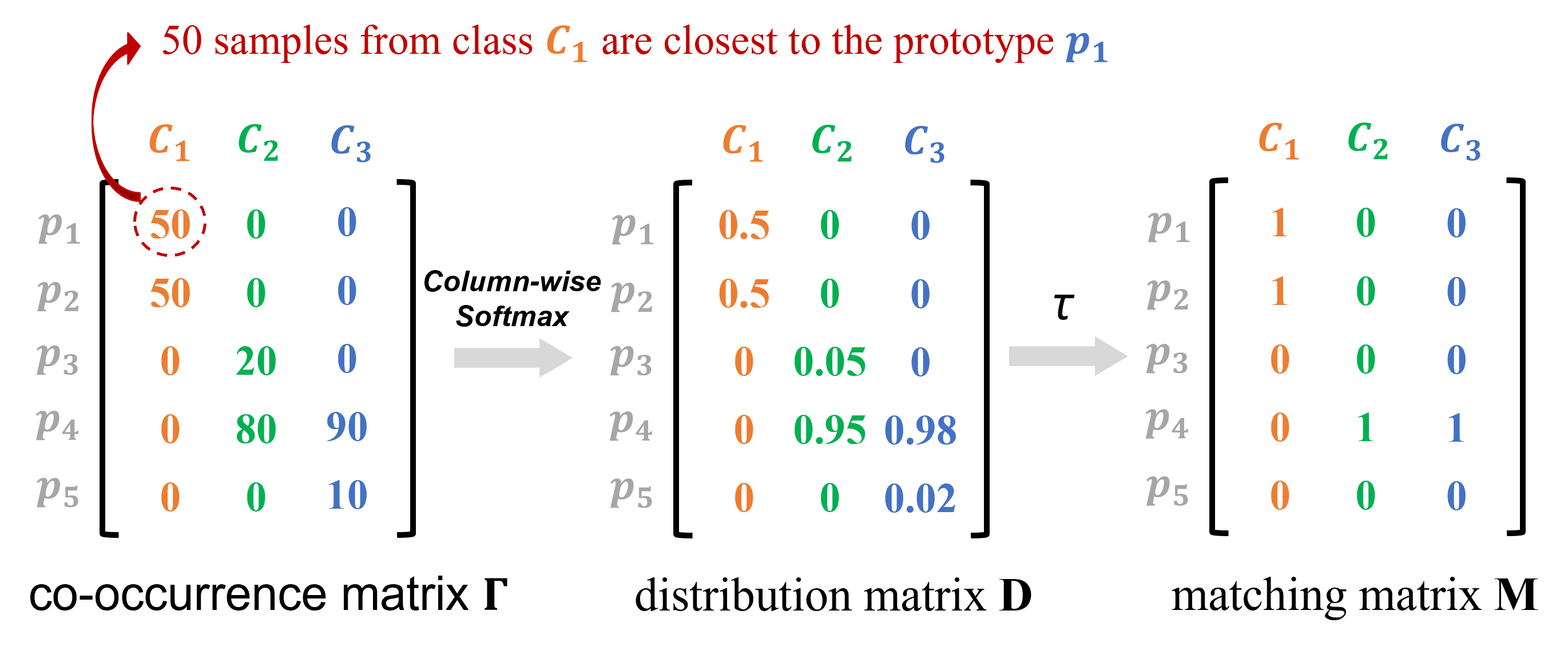}
    \vspace{-6mm}
    \caption{\textbf{The process of matching.} We first obtain the co-occurrence matrix $\Gamma$ between target prototypes and seen classes. Then, we apply a column-wise softmax to the co-occurrence matrix $\Gamma$ to get the distribution matrix $D$. Finally, we apply a threshold $\tau$ to each $D_{i,j}$ to obtain the matching matrix $M$. $M_{i,j} = 1$ means the class $C_j$ is matched to the prototype $p_i$.}
    \label{fig:match}
\end{figure}

We illustrate the matching procedure in Figure~\ref{fig:match}. In the example, samples in seen class $C_1$ are over-clustered into two clusters represented by the target prototypes $p_1$ and $p_2$, while samples in seen classes $C_2$ and $C_3$ are under-clustered and represented by only one target prototype $p_4$. After computing the matching matrix $M$, we match seen class $C_1$ to both target prototypes $p_1$ and $p_2$ and match seen classes $C_2$ and $C_3$ to target prototype $p_4$. Based on the result of matching, we consider $p_3$ and $p_5$, the unmatched target prototypes, to be the unseen prototypes.

\vspace{-2mm}
\vspace{-2mm}

\textbf{Threshold parameter $\tau$.}
$\tau$ is the threshold applied to each element in the distribution matrix $D$ (Equation \ref{equ:match}) to determine if there is a match between a target cluster and a seen class. We chose the threshold $\tau$ by observing the distribution matrix $D$. From our observations across all experiment settings, most elements in the distribution matrix $D$ are smaller than 0.02 (unmatched) or larger than 0.98 (matched), and a few elements are around 0.5 (matched, but over-clustering occurs).
Based on this observation, we choose $\tau = 0.3$ across all the experiments.
Note that the distribution matrix $D$ is obtained during training, so we do not use the label of the target samples (test set) to set this threshold.

\subsection{Fine-tuning step}
\label{sec:fine-tuning}

After the \textit{cluster-then-match} procedure, we initialize a linear classifier $W$ using the seen and unseen prototypes on top of the feature extractor $f_{\theta}$. In the final step of \method, we fine-tune both the feature extractor $f_{\theta}$ and the classifier $W$ to further improve model performance by updating the representation space and the prototypes. This step is illustrated in Figure~\ref{fig:concept} \textit{(iv)}.

In particular, a cross-entropy loss $L_{s}$~\cite{Lce} is applied to the source samples $x^s\in X^s$. This loss is used to transfer the knowledge of seen classes from the source samples to the target samples, and it is used to update the feature extractor and the seen prototypes:

\vspace{-4mm}
\begin{equation}
\label{equ:CE}
L_{s}(f_{\theta}, W_{seen}) = \frac{1}{N_s} \sum_{x^s\in X^s} - y(x^s) \log(p(y|x^s)),
\end{equation}

where $N_s$ is the number of source samples, $y(x)$ is the one-hot ground truth label of $x$, and $p(y|x) = \sigma(W^T\cdot f_{\theta}(x))$.

In addition, to balance the predictions of seen and unseen classes, we apply the regularization loss $L_{reg}$~\cite{reg_loss, orca} to the target samples $x^t\in X^t$. This term maximizes the entropy of the average of all the predictions:

\vspace{-4mm}
\begin{equation}
L_{reg}(f_{\theta}, W) = \frac{1}{N_t}\sum_{x^t\in X^t} p(y|x^t)\log(\frac{1}{N_t}\sum_{x^t\in X^t} p(y|x^t)),
\end{equation}

where $N_t$ is the number of target samples.

The final objective function in the fine-tuning step of the \method is as follows:

\begin{equation}
\label{equ:total}
    \min_{\theta, W} L_{s}(f_{\theta}, W_{seen}) + \lambda L_{reg}(f_{\theta}, W) ,
\end{equation}

where $\lambda$ denotes a regularization hyperparameter.

\section{Experiments}
\label{sec:experiment}
\begin{table*}[t]
\vskip -0.1in
\caption{Average H-score (\%) comparison of different seen/unseen splits on dataset Office, OfficeHome, VisDA, and DomainNet. We color the best and second-best results in red and blue.}
\label{table:main1}
\vspace{-4mm}
\begin{center}
\resizebox{\textwidth}{!}{
\begin{small}
\begin{sc}
\begin{tabular}{l|ccc|ccc|ccc|ccc|c}
\toprule
& \multicolumn{3}{c|}{Office} & \multicolumn{3}{c|}{OfficeHome} & \multicolumn{3}{c|}{VisDA} & \multicolumn{3}{c|}{DomainNet} & Average\\
seen/unseen& 21/10 & 16/15 & 10/21 & 45/20 & 33/32 & 20/45 & 8/4 & 6/6 & 4/8 & 240/105 & 173/172 & 105/240 &\\
\midrule
SIMPLE    & 64.9	&66.8	&78.3	&62.3	&66.0	&65.4	&55.4&	50.8&	50.9&	53.2	&55.9	&57.8&	60.6  \\
GCD    & 62.6 & 58.5 & 58.0 & 48.7 & 47.5 & 48.1 & 31.2 & 32.5 & 25.2 & 35.0 & 41.3 & 41.3 & 44.2  \\
ORCA    & 57.9 & 63.4 & 62.3 & 48.9 & 48.9 & 49.9 & 31.3 & 33.6 & 33.9 & 28.9 & 31.5 & 33.7 & 43.7\\
DCC    & 72.8 & 74.9 & 75.2 & 63.6 & 65.0 & 64.7 & 60.7 & 57.3 & 56.7 & 45.5 & 47.5 & 47.7 & 61.0  \\
DANCE    & 75.4 & 68.9 & 69.9 & 65.7 & 65.6 & 67.1 & 57.2 & 51.5 & 48.4 & \cellcolor{blue!20}56.3 & 55.7 & \cellcolor{blue!20}58.8 & 61.7 \\
OVAnet    & 73.2 & 75.7 & 75.3 & \cellcolor{blue!20}66.4 & \cellcolor{blue!20}68.6 & 68.7 & 59.9 & 60.8 & 60.2 & 54.2 & 55.6 & 58.5 & 64.7\\
UniOT    & 76.1 & 79.6 & \cellcolor{blue!20}83.4 & 64.4 & 64.9 & 64.8 & \cellcolor{blue!20}62.0 & 62.4 & 59.8 & 45.7 & 51.2 & 50.8 & 63.9 \\
NCDDA    & 80.3	&\cellcolor{blue!20}81.2	&81.7&	63.2&	64.3&	65.0&	57.2	&60.7	&59.3&	50.1&	52.6&	55.7	&64.3  \\
SAN    & \cellcolor{blue!20}80.5	&80.2	&82.0	&64.3&	65.0&	67.2&	61.2&	63.5&	61.5	&53.2&	54.8	&55.0&	65.7  \\
GLC    & 75.7 & 74.6 & 77.3 & 65.2 & 68.2 & \cellcolor{blue!20}69.3 & 61.2 & \cellcolor{blue!20}65.2 & \cellcolor{blue!20}62.7 & 54.9 & \cellcolor{blue!20}56.1 & 55.7 & \cellcolor{blue!20}65.8  \\
\textbf{\method}    & \cellcolor{red!20}84.7 & \cellcolor{red!20}84.9 & \cellcolor{red!20}85.6 &\cellcolor{red!20}69.4 & \cellcolor{red!20}69.6 & \cellcolor{red!20}70.2&\cellcolor{red!20}70.5 & \cellcolor{red!20}69.2 & \cellcolor{red!20}71.1 & \cellcolor{red!20}57.8 & \cellcolor{red!20}59.0 & \cellcolor{red!20}61.5 & \cellcolor{red!20}71.1\\
\bottomrule
\end{tabular}
\end{sc}
\end{small}
}
\end{center}
\vskip -0.2in
\end{table*}

\begin{table*}[t]
\caption{Average seen accuracy (\%), unseen accuracy (\%), and H-score (\%) of 50\% seen/unseen splits on dataset Office, OfficeHome, VisDA, and DomainNet. We color the best and second-best results in red and blue.}
\vspace{-4mm}
\label{table:main2}
\begin{center}
\resizebox{\textwidth}{!}{
\begin{small}
\begin{sc}
\begin{tabular}{l|ccc|ccc|ccc|ccc}
\toprule
& \multicolumn{3}{c|}{Office (16/15)} & \multicolumn{3}{c|}{OfficeHome (33/32)} & \multicolumn{3}{c|}{VisDA (6/6)} & \multicolumn{3}{c}{DomainNet (173/172)} \\
& seen & unseen & H-score & seen & unseen & H-score & seen & unseen & H-score & seen & unseen & H-score \\
\midrule
SIMPLE    & 62.3	&72.0	&66.8&	69.2	&63.1	&66.0&68.1	&40.5&	50.8&	69.1&	47.0&	55.9  \\
GCD    & 54.1 & 63.8 & 58.5 & 46.6 & 48.5 & 47.5 & 43.8 & 25.8 & 32.5 & 42.3 & 40.3 & 41.3  \\
ORCA    & 69.6 & 58.2 & 63.4 & 67.1 & 38.4 & 48.9 & 68.3 & 22.3 & 33.6 & 62.3 & 21.1 & 31.5 \\
DCC    & 78.0 & 72.1 & 74.9 & 70.3 & 60.5 & 65.0 & 75.3 & 46.2 & 57.3 & 50.2 & 45.1 & 47.5 \\
DANCE    & 73.3 &65.1  & 68.9 & \cellcolor{blue!20}72.0 & 60.3 & 65.6 & 70.2 & 40.7 & 51.5 & \cellcolor{blue!20}69.4 & 46.5 & 55.7 \\
OVAnet    & 76.7 & 74.8 & 75.7 & 71.8 & \cellcolor{blue!20}65.6 & \cellcolor{blue!20}68.6 & 60.4 & \cellcolor{blue!20}61.2 & 60.8 & 65.1 & 48.5 & 55.6 \\
UniOT    & 81.7 & \cellcolor{blue!20}77.6 & 79.6 & 70.5 & 60.2 & 64.9 & \cellcolor{blue!20}75.7 & 49.4 & 59.8 & 59.2 & 45.1 & 51.2 \\
NCDDA    & 88.9	&74.7	&\cellcolor{blue!20}81.2	&71.2&	58.6&	64.3&	70.4&	53.3	&60.7	&68.9&	42.5&	52.6 \\
SAN    & \cellcolor{blue!20}89.4	&72.7	&80.2	&\cellcolor{blue!20}72.0&	59.2&	65.0	&74.5	&55.3	&63.5&	67.3	&46.2&	54.8 \\
GLC    & 87.8 & 64.8 & 74.6 & \cellcolor{red!20}73.3 & 63.8 & 68.2 & 73.4 & 58.7 & \cellcolor{blue!20}65.2 & 62.9 & \cellcolor{blue!20}50.6 & \cellcolor{blue!20}56.1 \\
\textbf{\method}    & \cellcolor{red!20}{90.0} & \cellcolor{red!20}80.3 & \cellcolor{red!20}84.9 & 71.9 & \cellcolor{red!20}67.4 & \cellcolor{red!20}69.6 & \cellcolor{red!20}77.0& \cellcolor{red!20}62.8& \cellcolor{red!20}69.2 &\cellcolor{red!20}70.3 & \cellcolor{red!20}50.9 & \cellcolor{red!20}59.0 \\
\bottomrule
\end{tabular}
\end{sc}
\end{small}
}
\end{center}
\vskip -0.2in
\end{table*}

\subsection{Experimental setup}
\label{exp:setup}

\paragraph{Datasets.}
Universal domain adaptation (UniDA) shares the same assumption on data as cross-domain open-world discovery. Thus, we evaluate our method and the baselines on the standard UniDA benchmark datasets.
The \textbf{Office}~\cite{office} dataset has 31 classes and three domains: Amazon (A), DSLR (D), and Webcam (W). There are around 3K images in domain A and 1K in domains D and W. The \textbf{OfficeHome}~\cite{oh} dataset comprises 65 classes and four domains:  Art (A), Clipart (C), Product (P), and Real-World (R). There are around 4K images in domains C, P, and R, and 2K images in domain A. \textbf{VisDA}~\cite{visda} is a synthetic-to-real (S2R) dataset with 12 classes. There are around 150K images in domain S and 50K in domain R. \textbf{DomainNet}~\cite{domainnet} is the largest dataset, including 345 classes and six domains. Following the previous works~\cite{DN3,ovanet,uniot}, we use three domains: Painting (P), Real (R), and Sketch (S).

For each experimental setting, we create a pair of domains from one dataset, designating one domain as the source and another one as the target. Samples from the source domain have labels, while those from the target domain remain unlabeled.
Following the previous UniDA works~\cite{dance,ovanet,uniot}, we sort all the classes alphabetically and define the last $n$ class as unseen classes.
Then, we remove samples of the predefined unseen classes from the source set.
We evaluate \method and the baselines with different ratios of seen/unseen classes, including $70\%$, $50\%$, and $30\%$.

\paragraph{Evaluation metric.}
Open-world semi-supervised learning (OW-SSL) and cross-domain open-world discovery settings share the same task of recognizing seen and discovering unseen classes. Therefore, in line with the evaluation metric of the OW-SSL setting, we test the accuracy of both seen and unseen classes, referred to as seen and unseen accuracy.  
To compute unseen accuracy, we use the Hungarian algorithm~\cite{hungarian} to match the unseen classes and subsequently calculate the accuracy.

To evaluate the overall performance, we calculate the H-score~\cite{DN3}, as it provides a balanced measure of the performance of seen and unseen classes:
$$
H\_score = \frac{2 \cdot acc_{seen} \cdot acc_{unseen}}{acc_{seen} + acc_{unseen}}
$$

\paragraph{Baselines.}
We compare \method to UniDA and OW-SSL baselines as their settings are the closest to cross-domain open-world discovery. Since UniDA methods cannot discover novel classes, we extend them by first applying a UniDA method and then clustering the detected unseen samples to discover novel classes. OW-SSL methods do not need to be extended since they perform the same task even if under different assumptions about the data. We include as baselines two OW-SSL methods, namely ORCA~\cite{orca} and GCD~\cite{GCD}. We additionally compare to the six UniDA methods, namely DCC~\cite{dcc}, DANCE~\cite{dance}, OVANet~\cite{ovanet},  UniOT~\cite{uniot}, SAN~\cite{rebuttal2} and GLC~\cite{glc}. Also, we compare to NCDDA~\cite{rebuttal1}, which considers the cross-domain open-world discovery setting.

In addition, we design a simple baseline using the \textit{match-then-cluster} approach, referred to as SIMPLE.  
SIMPLE first trains the classifier on the source set with cross-entropy loss. Then, it predicts the labels and computes the prediction entropy for target samples. Samples with entropy exceeding a predefined threshold are considered unseen and undergo clustering. After labeling all the samples from seen and unseen classes, we finetune SIMPLE using the same objective function (\ref{equ:total}) proposed in CROW. More details are provided in Appendix~\ref{sec:simple}.

\paragraph{Implementation details.}
We use CLIP~\cite{clip} ViT-L~\cite{vit} as the feature extractor for \method and all the baselines. When fine-tuning, we update only the last two blocks in ViT-L and freeze the other parts following \citet{unida_foundation}, which shows that fine-tuning the whole ViT-L hurts the performance of the foundation model. More details are provided in the Appendix~\ref{apd:setting}. Our code is publicly available\footnote{\url{https://github.com/mlbio-epfl/crow}}.

UniDA methods are sensitive to the threshold used to separate seen and unseen samples. This threshold is crucial to the balance of seen and unseen accuracy. However, after changing the backbone from the the ImageNet pretrained ResNet50 to the CLIP ViT-L, the original threshold $\tau$ suggested in their works can lead to accuracy bias towards seen or unseen classes, resulting in a low H-score. To improve UniDA baselines and find optimal threshold $\tau$ that results in balanced results, we adapt the threshold \textit{using the test set}. This leads to an unrealistic evaluation setting since, in reality, we cannot use the test set to decide on the threshold, but our goal is to push the limits of the baselines in this setting. With the CLIP ViT-L backbone, our results substantially exceed the performance of all the baselines compared to their respective papers. We show the threshold $\tau$ we use for the baselines in the Appendix~\ref{apd:thre}. In contrast, CROW uses the same $\tau = 0.3$ and $\lambda = 0.1$ across all the experiments, and as we later show, it is robust to this threshold. 

\subsection{Results}
\label{sec:results}
\paragraph{Evaluation on benchmark datasets.}
We report the average H-score across four benchmark datasets: Office31, OfficeHome, VisDA, and DomainNet. We compare CROW to baselines with different ratios of seen and unseen classes, including $70\%$, $50\%$, and $30\%$ seen/unseen splits.
Table \ref{table:main1} shows that \method consistently outperforms all baselines in terms of H-score. In particular, across all datasets, \method achieves an $8\%$ relative improvement in the average H-score over the baselines. The detailed results of the $75$ different experimental settings with different pairs of source/target datasets are shown in Appendix~\ref{apd:more1}.

We next compare the performance separately on seen and unseen classes using the $50\%$ seen/unseen split. The results in Table \ref{table:main2} show that \method
consistently outperforms the baselines in discovering novel classes, achieving an $8.3\%$ average improvement over the baselines. On seen classes, \method outperforms baselines by a $2.9\%$ average improvement across all datasets. 
We observe similar results with $70\%$ and $30\%$ seen/unseen splits (Appendix~\ref{apd:more1}).

In comparison to UniDA methods that adopt the match-then-cluster strategy  (SIMPLE, DANCE, OVANet, UniOT, SAN, and GLC), \method outperforms the best baseline by $5.3\%$ in average H-score, highlighting the benefits of our cluster-then-match strategy.
When compared to the DCC, which also adopts a cluster-then-match strategy but follows a one-to-one matching procedure, \method outperforms DCC by $9.4\%$ in H-score. This underscores the benefits of the robust matching procedure.
Compared to OW-SSL methods (ORCA and GCD), we observe nearly $30\%$ improvement in average H-scores, indicating that OW-SSL methods cannot effectively overcome domain shifts and be applied in this setting.
Furthermore, \method surpasses NCDDA in H-score by 6.8\%, demonstrating its superior effectiveness in the cross-domain open-world discovery setting

In addition, we compare CROW to directly applying K-means to the CLIP features on the target datasets. We also compare our method to the CLIP zero-shot learning~\cite{clip}. The results show that our method outperforms these two methods by a large margin. We present the results and analysis in the Appendix~\ref{apd:k-means} and \ref{apd:clip-zeroshot}.

\begin{figure}[t]
    \centering
    \includegraphics[width=\linewidth]{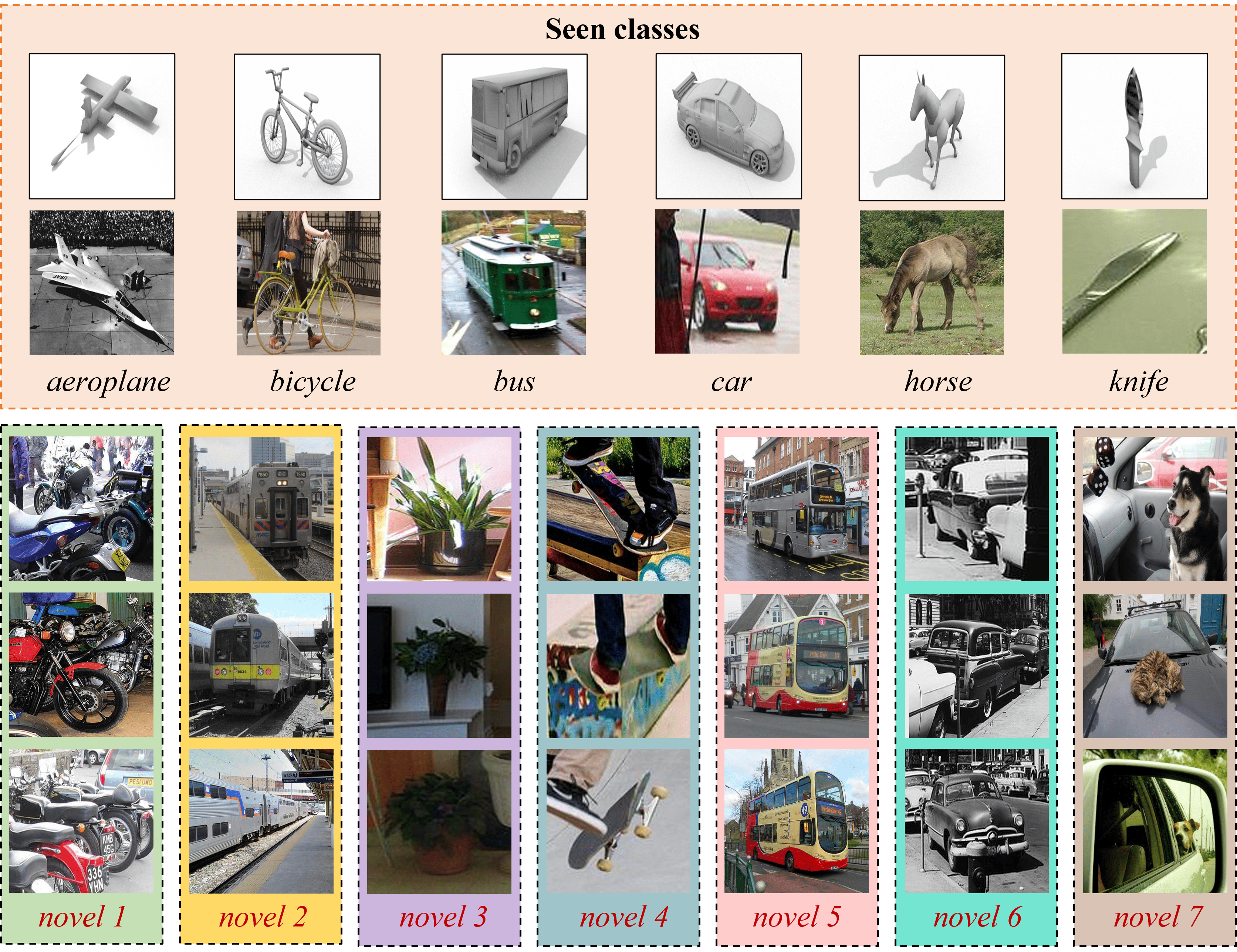}
    \vspace{-20pt}
    \caption{\textbf{Confident samples for seen and unseen classes} on VisDA. The synthetic images are from the source, and the real-world images are from the target. }
    \label{fig:top3} \vspace{-10pt}
\end{figure} 

\begin{figure*}[t]
\begin{center}
\centerline{\includegraphics[width=\linewidth]{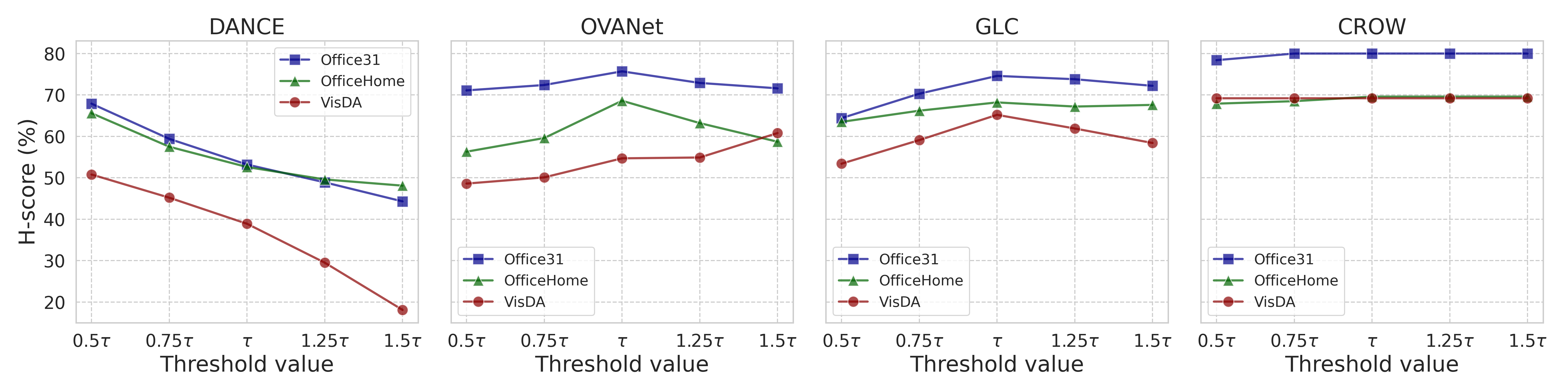}}
\vskip -0.2in
\caption{\textbf{Sensitivity to the threshold.} $\tau$ is the original threshold provided by our method and the previous works. We modify $\tau$ by scaling it with a multiplication factor.}
\label{fig:sensitivity}
\end{center}
\vskip -0.3in
\end{figure*}

\subsection{Qualitative results}

We visually inspect classes discovered by \method on the VisDA dataset.
Figure~\ref{fig:top3} shows the top-1 confident sample for each seen category and the top-3 confident samples for each novel category.  The results reveal that, in addition to annotating seen classes, \method successfully discovers seven unseen classes.
Notably, \method accurately discovers the VisDA predefined classes `\textit{motorcycle}', `\textit{plant}', `\textit{skateboard}', and `\textit{train}', which are absent in the source set. 
Additionally, the model recognizes double-decker buses, old-style cars, and cars with animals as novel classes. Despite discrepancies from the ground truth annotations (\textit{e.g.}, cars with animals are originally labeled as cars), the classes discovered by \method are meaningful. We further look at the VisDA predefined classes `\textit{person}' and `\textit{truck}' that are not discovered by \method. 
We find that confident predictions are people with skateboards that are classified into a novel class that corresponds to `\textit{skateboard}', again showing that the CROW's predictions are indeed meaningful. Furthermore, ground-truth class `\textit{truck}' is typically assigned to `\textit{car}', which is reasonable given many shared features. 
We elaborate more and show examples of failure cases in Appendix~\ref{apd:vis}.
Overall, this opens interesting research directions for designing proper evaluation strategies in this challenging setting since disagreement with the ground-truth annotations may not necessarily mean that the results are wrong, and even human annotators could disagree in these failure cases. 

\subsection{Ablation studies}

\paragraph{Benefits of fine-tuning.}
We evaluate how much fine-tuning helps to improve the performance of \method on the VisDA datataset. We compare \method in three settings: \textit{(i)} without fine-tuning (\textit{i.e.}, only clustering and matching steps), \textit{(ii)} with fine-tuning only the linear classifier $W$, and \textit{(iii)} with fine-tuning also the feature extractor of a foundation model. The results in Table \ref{table:freeze-finetune} show that fine-tuning both the feature extractor $f_\theta$ and the classifier $W$ helps to improve the performance. However, \method can still achieve high performance even without fine-tuning, by adopting only clustering and matching steps.

\vspace{-3mm}
\begin{table}[h]
\caption{Seen, unseen accuracy (\%) and H-score (\%) of our method with different fine-tuning strategies on the VisDA dataset (6/6).} \vspace{-10pt}
\label{table:freeze-finetune}
\begin{center}
\resizebox{\linewidth}{!}{
\begin{small}
\begin{sc}
\begin{tabular}{lccc}
\toprule
 & Seen & Unseen & H-score \\
\midrule
Without Fine-tune & 73.8& 61.2& 66.9 \\
Fine-tune only $W$  & 75.2& 61.8& 67.8 \\
Fine-tune $f_{\theta}$ and $W$ & 77.0& 62.8& 69.2 \\
\bottomrule
\end{tabular}
\end{sc}
\end{small}
}
\end{center}
\vskip -0.2in
\end{table}

\paragraph{Sensitivity to threshold $\tau$.}
We next evaluate the sensitivity to the thresholds of CROW and UniDA baselines on the $50\%$ seen/unseen split of the Office31, OfficeHome, and VisDA datasets. We compare \method, DANCE, OVANet, and GLC across different values of their respective threshold. 
CROW has a matching threshold $\tau$ (Equation~\ref{equ:match}). 
DANCE has a threshold that detects unseen samples using the entropy of the prediction. OVANet and GLC have a threshold that detects unseen samples using the prediction confidence. Due to the different scales of the thresholds employed by each method, we test values from $0.5 \tau$ to $1.5 \tau$, where $\tau$ denotes the default threshold used in these previous works. We evaluate the effect of changing the threshold on performance in Figure~\ref{fig:sensitivity}. The results show that \method is extremely robust to the threshold variations. However, this is not the case for baseline methods.
For example, DANCE demonstrates considerable sensitivity to threshold changes, and the original $\tau$ deviates from the optimal value after changing the backbone. For OVANet and GLC, their original $\tau$ yields good performance in the majority of cases, but these methods still exhibit sensitivity to the threshold.

\paragraph{Ablation study on the objective function.}
To investigate the importance of each part of the objective function in Equation~\ref{equ:total}, we conduct an ablation study on the VisDA dataset.
Table~\ref{table:ablation} shows that the removal of the supervised loss $L_s$ results in a decrease in seen accuracy, while the absence of the entropy regularization $L_{reg}$ causes the accuracy to bias toward seen classes. The best performance is achieved when combining the two losses.

\vspace{-2mm}
\begin{table}[h]
\caption{Ablation study on the objective function. We show the seen/unseen accuracy, and H-score (\%) on the VisDA dataset (6/6).} 
\label{table:ablation}
\begin{center}
\resizebox{0.8\linewidth}{!}{
\begin{small}
\begin{sc}
\begin{tabular}{lccc}
\toprule
Approach & Seen & Unseen & H-score \\
\midrule
w/o $L_s$   & 65.6& 61.5& 63.5 \\
w/o $L_{reg}$ & 77.2& 56.9& 65.7 \\
CROW & 77.0& 62.8& 69.2 \\
\bottomrule
\end{tabular}
\end{sc}
\end{small}
}
\end{center}
\vskip -0.2in
\end{table}

\paragraph{\method with different foundation models.} In all experiments, we used CLIP as the feature extractor. We next compare the performance of \method in the space of different foundation models. As foundation models, we use CLIP~\cite{clip}, DINO\_v2~\cite{dinov2}, and SWAG~\cite{swag} across varying sizes of the ViT. Table \ref{table:fundation} illustrates that \method achieves better performance with stronger feature extractors. This suggests that \method can benefit from further advancement in the field by using stronger foundation models as a feature extractor.

\vskip -0.1in
\begin{table}[h]
\caption{Seen, unseen accuracy (\%) and H-score (\%) of CROW with different pretrained foundation models on VisDA (6/6).} \vspace{-10pt}
\label{table:fundation}
\begin{center}
\resizebox{\linewidth}{!}{
\begin{small}
\begin{sc}
\begin{tabular}{lcccc}
\toprule
Method&Backbone & Seen & Unseen & H-score \\
\midrule
CLIP  & ViT-B &74.5& 58.9& 65.8 \\
  & ViT-L &77.0& 62.8& 69.2 \\
  \midrule
DINO\_v2 & ViT-B & 74.2& 55.5& 63.5 \\
 & ViT-L & 76.8& 57.6& 65.8 \\
 & ViT-G & 78.2& 60.4& 68.2\\
 \midrule
SWAG & ViT-B & 74.8& 60.2&66.7\\
 & ViT-L & 78.4& 63.0& 69.9 \\
 & ViT-H & 79.0& 63.4& 70.3 \\
\bottomrule
\end{tabular}
\end{sc}
\end{small}
}
\end{center}
\end{table}

Results of \method with the estimated number of novel classes $|C_t|$ and on the UniDA data split are shown in Appendix~\ref{apd:estimated-c} and ~\ref{apd:unida-split}.

\subsection{Pretrained model for unseen classes}

A potential problem is that the pretrained feature extractors might have encountered the unseen classes during pretraining. Indeed, this is a common issue in the research fields of novelty detection and category discovery, and it existed even before the age of foundation models. For example, most previous works on open-set/universal domain adaptation, novel class discovery, and general class discovery~\cite{ovanet,dance,dcc,GCD} use ImageNet pretrained ResNet-50 as their backbone, and some of the unseen classes are present in the ImageNet dataset.

To test whether a model that has seen instances of unseen classes can artificially inflate the results, we perform an experiment on the Office31 dataset in which we train two versions of the ResNet-50 feature extractor in a supervised fashion: (1) trained on the whole ImageNet dataset, and (2) trained on the ImageNet dataset without the samples of the unseen classes (\textit{e.g.}, we remove samples of `\textit{desk}', `\textit{barber chair}', `\textit{folding chair}', `\textit{rocking chair}’ from the ImageNet dataset for the unseen class `\textit{desk\_chair}’). Table~\ref{table:unseensample} shows that whether the model has seen instances of unseen classes only marginally affects performance. Furthermore, it is important to emphasize that the model trained without instances of unseen classes has also seen fewer different samples and less data in general, so we cannot fully attribute these small differences to the fact that the model has seen unseen samples. However, how to avoid this common problem in the research fields of novelty detection and category discovery still needs to be further explored~\cite{rebuttal_discussion}.

\vskip -0.1in
\begin{table}[h]
\caption{Ablation study on different pretrained datasets. We show the seen, unseen accuracy (\%) on the Office dataset.} 
\label{table:unseensample}
\begin{center}
\resizebox{0.8\linewidth}{!}{
\begin{small}
\begin{sc}
\begin{tabular}{lcc}
\toprule
Dataset & Seen & Unseen\\
\midrule
whole ImageNet   & 79.5& 56.4 \\
ImageNet w.o. unseen  & 79.6& 55.8 \\
\bottomrule
\end{tabular}
\end{sc}
\end{small}
}
\end{center}
\vskip -0.2in
\end{table}

\section{Limitations}
\label{sec:limitation}
In \method, the ability to discover novel classes heavily relies on clustering within the representation space established by the foundation models. Consequently, CROW may exhibit worse performance on datasets where the foundation models lack robustness. For example, we evaluated our method on DomainNet, from Sketch (source) to Quickdraw (target). Quickdraw contains images with grey-level lineart, and CLIP is not robust to that image style. Under the setting of the $50\%$ seen/unseen split and the exact same training setup as described in the Implementation details in Section~\ref{exp:setup}, CROW achieves only $20.4\%$ seen accuracy and $24.2\%$ unseen accuracy on this dataset. However, the strongest baseline GLC achieves even worse results with $20.1\%$ seen accuracy and $18.6\%$ unseen accuracy. These results indicate that further exploration needs to be done to deal with challenging datasets like DomainNet Quickdraw.

\section{Conclusion}
\label{sec:conclusion}
In this work, we address the gap between open-world semi-supervised learning and universal domain adaptation by considering a cross-domain open-world discovery setting that encompasses both categorical and distributional shifts. To tackle this challenging problem, we propose \method, a prototype-based method built upon foundation models. 
CROW combines source seen prototypes and target unseen prototypes through a robust \textit{cluster-then-match} approach, simultaneously accomplishing seen class recognition and unseen class discovery.
By conducting experiments across $75$ different categorical-shift and domain-shift situations, we demonstrate that \method consistently outperforms alternative baselines and effectively overcomes the challenges of the cross-domain open-world discovery setting.

\clearpage

\section*{Acknowledgements}
The authors thank Artyom Gadetsky, Liangze Jiang, Matej Grcić, and Ramon Vinas Torné, for their feedback on our manuscript. We also thank Chanakya Ekbote and Yulun Jiang for discussing this work.  We gratefully acknowledge the support of EPFL and ZEISS.

\section*{Impact Statement}
This paper presents work whose goal is to advance the field of Machine Learning. There are many potential societal consequences of our work, none which we feel must be specifically highlighted here.

\bibliography{main}
\bibliographystyle{icml2024}

\clearpage

\appendix


\section{Implementation details}
\subsection{Training details}
\label{apd:setting}
Our core algorithm is developed using PyTorch~\cite{pytorch}. We use CLIP ViT-L14-336px as the backbone for all the methods. When fine-tuning, we update only the last two blocks of CLIP ViT-L14-336px and freeze the other parts. For the classifier, CROW uses the normalized linear classifier as described in Section~\ref{sec:overview}. For the baselines, we use the same classifier architecture originally proposed by their works, and we only change the input dimension of the classifiers to match the feature dimension.

For optimizing, we use the SGD optimizer for all experiments, and the learning rate is set to 0.001 for the classifier and 0.0001 for the feature extractor (CLIP ViT-L14-336px). We set the batch size to 32 and train all the methods for 1K iterations. Since there is no validation set in our setting, we report the results of the last iteration. 

\subsection{Implementation details about baselines}
\label{sec:simple}

We directly apply the OW-SSL methods GCD and ORCA and the method NCDDA to our experiment settings. However, for the other baselines, we adapt them to address our problem setting. The detailed procedures for adaptation are outlined below.

\paragraph{SIMPLE.} 
SIMPLE shares the same network architecture as CROW (a feature extractor $f_{\theta}$ and a normalized linear classifier $W$), and it uses the match-then-cluster approach. Specifically, it first trains the classifier $W$ on the source set with cross-entropy loss (Equation~\ref{equ:CE}). Importantly, since SIMPLE trains the model only on the source set, it is likely to make the predictions biased to the seen classes. To prevent this, we freeze the feature extractor from SIMPLE.


After training the classifier, we predict the labels and compute the entropy for all the samples. Specifically, given input $x$, we first extract the feature by $ z = f_\theta (x) $.  
Then, we calculate the output vector $p(y|x)$ using $p(y|x) = \sigma(W^T\cdot z)$, where $\sigma$ is the softmax activate function. We predict the label using $c = \arg\max_i p^i$. Then, we calculate the entropy $H$ for the output vector $p(y|x)$. If $H$ is larger than a predefined threshold $\tau$, we assign this sample to be an unseen sample. Note that since DANCE also applies a threshold to the entropy of prediction, we use the $\rho$ in DANCE as the $\tau$ here.
After predicting labels for all samples, we cluster detected unseen samples using K-means with $K = |C_t| - |C_s|$ to discover novel classes. We assume the number of target classes $|C_t|$ is given as a prior.
After labeling all the samples from seen and unseen classes, we finetune SIMPLE using the same objective function \ref{equ:total} as in CROW.

\paragraph{UniDA methods (except DCC).}
For all the UniDA methods except DCC (DANCE, OVANet, UniOT, SAN, GLC), we use them as the match-then-cluster approach. Specifically, we first apply the methods to predict the labels of the target samples. Then, each target sample is labeled as a seen class or the class unseen. After labeling, we cluster all the samples labeled as class unseen using K-means with $K = |C_t| - |C_s|$ to discover novel classes. We assume the number of target classes $|C_t|$ is given as a prior.

\paragraph{DCC.}
Different from the other UniDA methods, we use DCC as a cluster-then-match approach. Thus, we follow the original steps of DCC. We change only one thing: in the original work, DCC estimates the number of target classes $|C_t|$, but we directly use $|C_t|$ as a prior in DCC for a fair comparison.



\subsection{Threshold adaptation}
\label{apd:thre}

As mentioned in Section~\ref{exp:setup}, we adapt the threshold $\tau$ for the baselines when needed. Table~\ref{table:hyper} shows how we change $\tau$ for the baseline methods to obtain balanced seen and unseen accuracy. $\tau$ is the original threshold provided by the previous works, and we scale it with a multiplication factor. ($\tau$ is the $\rho$ in DANCE and SIMPLE, 0.5 with no name in OVANet.)

\begin{table}[h]
\caption{Hyper-parameter changing. $\tau$ is the original threshold provided by the previous works.}
\vskip -0.1in
\label{table:hyper}
\begin{center}
\resizebox{\linewidth}{!}{
\begin{small}
\begin{sc}
\begin{tabular}{lcccc}
\toprule
 & Office & OfficeHome & VisDA & DomainNet \\
\midrule
SIMPLE & 0.3$\tau$& 0.5$\tau$& 0.3$\tau$& 0.7$\tau$ \\
DANCE  & 0.3$\tau$& 0.5$\tau$& 0.3$\tau$& 0.7$\tau$ \\
OVAnet  & -& -& 1.5$\tau$& - \\
\bottomrule
\end{tabular}
\end{sc}
\end{small}
}
\end{center}
\vskip -0.1in
\end{table}

\section{Additional results}
\label{apd:additional}

\subsection{Comparison to the K-means}\label{apd:k-means}
This section shows the results and analysis of comparing our method CROW to applying K-means to the CLIP features.

\begin{table}[h]
\caption{H-score (\%) comparison between K-means and CROW.} \vspace{-1pt}
\label{table:k-means}
\begin{center}
\resizebox{\linewidth}{!}{
\begin{small}
\begin{sc}
\begin{tabular}{lcccc}
\toprule
 & Office & OfficeHome & VisDA & DomainNet \\
\midrule
K-means  & 77.2 &65.9& 62.4& 52.7 \\
CROW  & 84.9 &69.9& 69.2& 59.0 \\
\bottomrule
\end{tabular}
\end{sc}
\end{small}
}
\end{center}
\end{table}

Table~\ref{table:k-means} shows that our method outperforms K-means by a large margin. Moreover, it is important to note that our method labels the seen classes while applying K-means to the CLIP features only separates different classes without matching the clusters to the seen classes, which means our H-score is tested on a harder task.

\begin{figure}[t]
    \centering
    \includegraphics[width=0.9\linewidth]{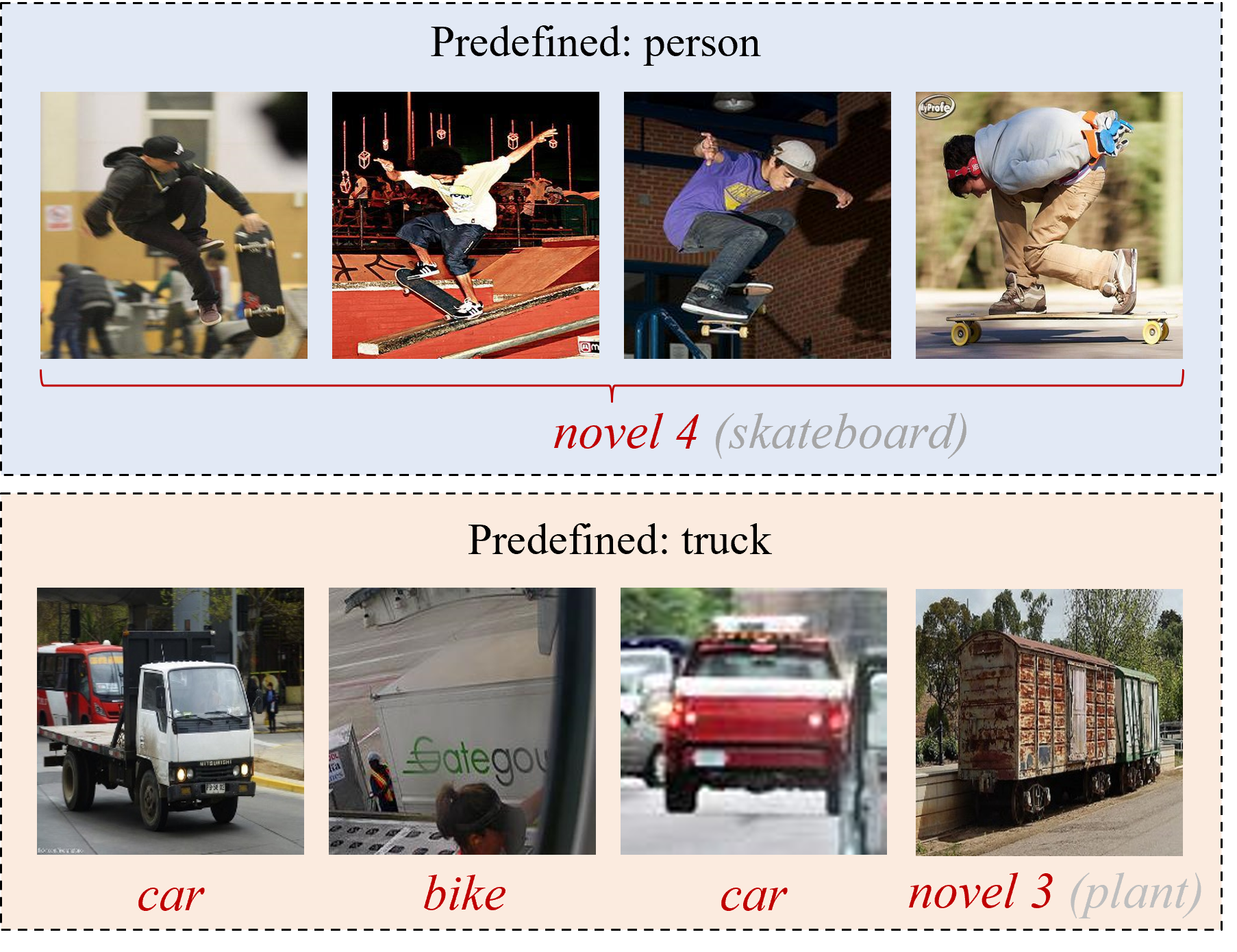}
    \caption{\textbf{Failure cases} on VisDA (6/6). The samples are the top 4 confident samples (top 1 to 4 from left to right) for class `person' and `truck'.}
    \label{fig:failure_cases}
\end{figure}

\subsection{Comparison to the CLIP zero-shot learning}\label{apd:clip-zeroshot}

Another simple way to address cross-domain open-world discovery is to apply CLIP zero-shot learning with a large vocabulary list.
Here, we show the results and analysis of comparing our method CROW to CLIP zero-shot learning. 

We do two experiments on the Office31 dataset. In the first experiment, the vocabulary list of CLIP contains only the names of the 31 classes. In the second experiment, we create a large vocabulary list for CLIP by combining the names of the 31 classes from the Office dataset and the 345 classes from the DomianNet dataset. We remove the names of the duplicate classes from the vocabulary list. We consider the first 16 classes from the Office31 dataset seen and the last 15 classes unseen. Table~\ref{table:clip-zeroshot} shows the results. The results show that our method achieves a comparable performance even if we provide CLIP with an exact ground truth vocabulary list, and our method outperforms CLIP by a large margin if CLIP uses a large-enough vocabulary list. 

\begin{table}[h]
\caption{H-score (\%) comparison between CLIP zero-shot learning and CROW.} \vspace{-1pt}
\label{table:clip-zeroshot}
\begin{center}
\resizebox{\linewidth}{!}{
\begin{small}
\begin{sc}
\begin{tabular}{lc}
\toprule
 & H-score \\
\midrule
CLIP zero-shot + 31 classes vocabulary list	  & 85.7 \\
CLIP zero-shot + large vocabulary list	& 75.4 \\
CROW  & 84.9  \\
\bottomrule
\end{tabular}
\end{sc}
\end{small}
}
\end{center}
\end{table}

\subsection{Failure cases}
\label{apd:vis}

In Figure~\ref{fig:top3}, we show that we successfully discover four predefined classes out of six. Figure~\ref{fig:failure_cases} shows the possible reason why we do not manage to discover the other two predefined classes `person' and `truck'. For the predefined class `person', we can see that the top four confident samples are all persons with skateboards, and they are classified into class `novel 4', which is `skateboard' as shown in Figure~\ref{fig:top3}. For the predefined class `truck', we can see that two of them are classified into `car', and this might be because they share lots of common features with cars; one is classified into `bike', a possible explanation is because of the green logo `Gate' is close to bike; one is classified into class `novel 3, which is `plant', possibly because of the full-of-tree background.

\subsection{Sensitivity to $\lambda$}
There is a hyper-parameter $\lambda$ in Equation~\ref{equ:total}, which stands for the weight for the regularization term.
Table~\ref{table:lambda} shows that our method is robust to $\lambda$ as long as it is not set to be extremely huge or tiny.

\begin{table}[h]
\caption{Seen accuracy (\%), unseen accuracy (\%), and H-score (\%) of \method with different $\lambda$ on the VisDA (6/6).}
\label{table:lambda}
\begin{center}
\begin{small}
\begin{sc}
\begin{tabular}{cccc}
\toprule
$\lambda$ & Seen & Unseen & H-score \\
\midrule
0.1 & 78.5& 62.0& 69.3 \\
0.3 & 78.8& 61.3& 69.0 \\
0.5 & 77.8& 61.7& 68.8 \\
0.7 & 77.6& 62.7& 69.4 \\
0.9 & 77.3& 62.6& 69.2 \\
1.0 & 77.0& 62.8& 69.2 \\
\bottomrule
\end{tabular}
\end{sc}
\end{small}
\end{center}
\vskip -0.1in
\end{table}

\subsection{\method with the estimated number of clusters} \label{apd:estimated-c}
In the clustering step, we assume we know the number of unseen classes $|C^t|$ following the OW-SSL setting. However, in practice, we sometimes do not know the real $|C^t|$. Under this condition, we need to estimate $|C^t|$.

We estimate $|C^t|$ using the technique proposed in \cite{estimateK}. In the original work, it estimates $|C^t|$ by applying K-means with different $K$ on both source and target samples. Then, it tests the cluster accuracy using Hungarian algorithm~\cite{hungarian} for the labeled samples and selects the $K$ that leads to the best cluster accuracy. However, since there is a domain shift between source and target set in our problem setting, we apply K-means only on the target data instead of both source and target data. Other steps remain the same. Table \ref{table:unknown} shows the results of using the real $|C^t|$ and the estimated $|C^t|$, and we can see that the results are still good with estimated $|C^t|$.

\begin{table}[h]
\caption{H-score of using estimated $|C^t|$. We show 50\% seen/unseen split as an example.}
\vskip -0.1in
\label{table:unknown}
\begin{center}
\resizebox{\linewidth}{!}{
\begin{small}
\begin{sc}
\begin{tabular}{lcccc}
\toprule
 & Office & OfficeHome & VisDA & DomainNet \\
\midrule
Known & 84.9& 69.6& 69.2& 59.0 \\
Estimated  & 81.5& 67.2& 68.0& 57.8 \\
\bottomrule
\end{tabular}
\end{sc}
\end{small}
}
\end{center}
\vskip -0.1in
\end{table}

\subsection{Evaluation on the UniDA data split.} \label{apd:unida-split}
Since UniDA is the closest setting to CD-OWD, we demonstrate the efficacy of our method on the UniDA data split. 

There are four possible relationships: closed set~\cite{UDA}, partial set~\cite{pda}, open set~\cite{osda1}, and open partial set~\cite{unida}.
The term \textit{partial} means there are classes that exist only in the source set.
UniDA setting considers all four possible relationships between source and target sets. Thus, we test our method on the four conditions of closed set, partial set, open set, and open partial set on the VisDA dataset. We use the same evaluation metric as the previous experiments.
Note that we assume we know the number of classes of the target sets but do not know the relationship between source and target sets. For example, in the condition of the closed set, we assume we know that there are 12 classes in the target set. However, we do not know if there are novel classes, so we will still detect unseen. 
Table \ref{table:UniDA} shows that our method achieves comparable performance with UniDA data split. 

\begin{table}[h]
\caption{Results of UniDA data split on VisDA. The numbers of (shared classes/source private classes/target private classes) for close-set, partial-set, open-set, and open-partial-set are 12/0/0, 6/6/0, 6/0/6, and 6/3/3. The results are H-score when the number of target private classes is not zero; otherwise, we show the accuracy of shared classes.}
\label{table:UniDA}
\begin{center}
\resizebox{\linewidth}{!}{
\begin{small}
\begin{sc}
\begin{tabular}{lcccc}
\toprule
 & Close & Partial & Open & Open-partial\\
\midrule
DCC & 80.1 & 79.8& 57.3& 65.2 \\
OVANet & 78.0 & 73.2& 60.8& 61.2 \\
GLC & 75.6 & 81.2& 65.2& 72.2 \\
\method & 79.9 & 80.4& 69.2& 73.4 \\
\bottomrule
\end{tabular}
\end{sc}
\end{small}
}
\end{center}
\vskip -0.1in
\end{table}
\vspace{-2mm}

\subsection{Detailed experimental results} \label{apd:more1}
Table~\ref{table:main1} only shows the average H-score on each dataset. 
Here, we show the detailed results of each categorical-shift and domain-shift scenario on different datasets in Table~\ref{table:office2110} to \ref{table:oh20}.
Each table shows the result of one dataset with one data split. For example, Office (21/10) shows the result of the Office dataset with a 21/10 seen/unseen data split. In each table, we show all the domain-shift scenarios. For example, there are three domains in the Office dataset, Amazon (A), DSLR (D), and Webcam (W). Then, there are six pairs: A2D, A2W, D2A, D2W, W2A, and W2d, where A2D means from Amazon (source set) to DSLR (target set).

Table~\ref{table:main2} only shows the average seen accuracy, unseen accuracy, and H-score on each dataset of 50\% seen/unseen split. Here, we show the average seen accuracy, unseen accuracy, and H-score on each dataset of 50\% seen/unseen split in Table~\ref{table:office2110} to \ref{table:oh20}.


\begin{table}[h]
\caption{H-score (\%) of dataset Office (21/10) on different pairs of domain. We color the best and second-best results in red and blue.}
\label{table:office2110}
\begin{center}
\resizebox{\linewidth}{!}{
\begin{small}
\begin{sc}
\begin{tabular}{l|ccccccc}
\toprule
& A2D & A2W & D2A & D2W & W2A & W2D & Avg. \\
\midrule
SIMPLE    &62.6	&60.8	&43.4	&86.2&	42.3&	90.3	&64.9 \\
GCD    & 62.6 & 58.5 & 58.0 & 48.7 & 47.5 & 48.1 & 31.2 \\
ORCA   & 65.9	&52.5	&51.5	&65.3	&54.8	&52.1	&57.9\\
DCC    & 66.3&	72.5	&69.8&	84.5&	66.9&76.8&	72.8 \\
DANCE    & 79.4	&\cellcolor{blue!20}77.5	&61.4&	82.9	&62.3	&87.8	&75.4  \\
OVAnet    & 77.4&	67.1&	59.8&	\cellcolor{blue!20}88.0&55.9	&90.7	&73.2\\
UniOT    & \cellcolor{blue!20}82.8	&70.7	&65.3	&78.1	&66.5	&\cellcolor{blue!20}91.5	&76.1 \\
NCDDA&81.2	&76.4&	\cellcolor{red!20}78.7&	80.8&	\cellcolor{blue!20}76.0&	87.1	&80.3\\
SAN&80.7	&76.7	&\cellcolor{blue!20}77.8&	81.4&	\cellcolor{red!20}80.6&	85.2&	\cellcolor{blue!20}80.5\\
GLC    & 78.5	&71.9	&73.2&	72.6	&73.4&	81.9	&75.5  \\
\method    &\cellcolor{red!20}87.9&	\cellcolor{red!20}90.3&	70.3	&\cellcolor{red!20}93.0	&70.2&	\cellcolor{red!20}92.8&	\cellcolor{red!20}84.7\\
\bottomrule
\end{tabular}
\end{sc}
\end{small}
}
\end{center}
\end{table}

\begin{table}[h]
\caption{H-score (\%) of dataset Office (16/15) on different pairs of domain. We color the best and second-best results in red and blue.}
\label{table:office1615}
\begin{center}
\resizebox{\linewidth}{!}{
\begin{small}
\begin{sc}
\begin{tabular}{l|ccccccc}
\toprule
& A2D & A2W & D2A & D2W & W2A & W2D & Avg. \\
\midrule
SIMPLE    &65.8	&57.6	&47.6	&84.8&	47.8&	\cellcolor{blue!20}90.6&	66.8\\
GCD    & 58.8&	37.0&	50.6	&76.6&	46.6&	69.9&	58.5 \\
ORCA    & 61.4	&67.4	&54.3&	80.3	&46.1&	65.9	&63.4\\
DCC     & 68.3	&72.5	&71.8	&85.5&	68.9&	77.8&	74.9 \\
DANCE    & 65.7	&67.3	&59.7&	83.4	&54.1&	82.7&	68.9  \\
OVAnet    & 81.6	&76.3	&61.5&	\cellcolor{blue!20}91.2	&51.2	&\cellcolor{blue!20}90.6	&75.7 \\
UniOT    & 79.6	&\cellcolor{blue!20}83.8	&72.7&	90.9	&75.0	&85.7	&\cellcolor{blue!20}81.5 \\
NCDDA&\cellcolor{blue!20}83.9&	76.4&	\cellcolor{red!20}80.0&	78.7&	\cellcolor{red!20}81.8&	85.5	&81.2\\
SAN&79.5	&76.0&	\cellcolor{blue!20}78.3&	80.3	&\cellcolor{blue!20}80.6	&85.5&	80.2\\
GLC    & \cellcolor{red!20}85.4	&74.2&	69.8	&79.1	&63.2	&71.3	&74.6  \\
\method    & 83.8&	\cellcolor{red!20}85.9	&73.9&	\cellcolor{red!20}94.8&	78.1&	\cellcolor{red!20}91.2	&\cellcolor{red!20}84.9\\
\bottomrule
\end{tabular}
\end{sc}
\end{small}
}
\end{center}
\end{table}

\begin{table}[h]
\caption{H-score (\%) of dataset Office (10/21) on different pairs of domain. We color the best and second-best results in red and blue.}
\label{table:office1021}
\begin{center}
\resizebox{\linewidth}{!}{
\begin{small}
\begin{sc}
\begin{tabular}{l|ccccccc}
\toprule
& A2D & A2W & D2A & D2W & W2A & W2D & Avg. \\
\midrule
SIMPLE    &76.7	&78.9	&71.9	&84.1	&71.1	&85.5	&78.3\\
GCD    & 53.0	&38.2	&51.0&	77.5	&40.9	&70.9	&58.0\\
ORCA    & 61.0	&57.3	&44.9	&75.0&	60.2	&69.0	&62.3 \\
DCC     & 69.3&	76.5&	74.8	&86.5	&69.5&	74.8&	75.2 \\
DANCE    & 77.6	&73.3	&58.2	&74.5	&56.8&	77.3	&69.9  \\
OVAnet    & \cellcolor{blue!20}84.8	&78.1	&50.7&	90.8&	48.6	&90.0&	75.3 \\
UniOT    & 84.1&	\cellcolor{red!20}84.5	&69.9&	\cellcolor{blue!20}90.9	&77.2	&\cellcolor{red!20}93.7	&\cellcolor{blue!20}83.4 \\
NCDDA&84.1	&75.0&	\cellcolor{blue!20}80.6&	80.3&	\cellcolor{red!20}82.3&	87.1&	81.7\\
SAN&80.8&	79.7&	\cellcolor{red!20}81.0&	83.7&	\cellcolor{blue!20}81.3&	85.6&	82.0\\
GLC    & 82.9	&77.5	&72.6	&77.3	&73.0	&78.5	&77.3  \\
\method    & \cellcolor{red!20}85.8	&\cellcolor{blue!20}82.9	&77.5	&\cellcolor{red!20}96.0&	79.1&	\cellcolor{blue!20}91.7&	\cellcolor{red!20}85.6\\
\bottomrule
\end{tabular}
\end{sc}
\end{small}
}
\end{center}
\end{table}

\begin{table}[h]
\caption{H-score (\%) of dataset DomainNet (240/105) on different pairs of domain. We color the best and second-best results in red and blue.}
\label{table:DN240}
\begin{center}
\resizebox{\linewidth}{!}{
\begin{small}
\begin{sc}
\begin{tabular}{l|ccccccc}
\toprule
& P2R & P2S & R2P & R2S & S2P & S2R & Avg. \\
\midrule
SIMPLE    & 58.0&	45.7	&52.2&	54.8&	48.9&	58.7	&53.2\\
GCD    & 39.4	&28.7	&33.6	&36.4	&34.1&	36.9	&35.0\\
ORCA    & 32.7	&20.4&	21.0	&33.3&	28.4&	36.0	&28.9\\
DCC    & 52.4&	42.6	&44.5&	44.1&	41.1&	47.3	&45.5 \\
DANCE    &61.4	&\cellcolor{red!20}52.3	&\cellcolor{red!20}55.6&	52.1&	\cellcolor{red!20}54.3&	61.5&	\cellcolor{blue!20}56.3\\
OVAnet    & 59.6	&49.1	&\cellcolor{blue!20}54.0	&49.9&	51.8	&59.2	&54.2\\
UniOT    & 49.1	&42.6	&45.8	&43.4&	42.8&	49.6&	45.7 \\
NCDDA&61.4&	47.9&	41.6&	45.8&	41.0&	61.3&	50.1\\
SAN& \cellcolor{blue!20}62.7&	51.2&	45.2&	\cellcolor{blue!20}52.5&	42.8&	63.3&	53.2\\
GLC    & 61.3&	51.5&	46.4	&\cellcolor{red!20}53.1	&48.0&	\cellcolor{blue!20}65.6	&54.9  \\
\method    & \cellcolor{red!20}68.0	&\cellcolor{red!20}52.3	&51.7&	49.2	&\cellcolor{blue!20}53.6	&\cellcolor{red!20}69.9&	\cellcolor{red!20}57.8\\
\bottomrule
\end{tabular}
\end{sc}
\end{small}
}
\end{center}
\end{table}

\begin{table}[h]
\caption{H-score (\%) of dataset DomainNet (173/172) on different pairs of domain. We color the best and second-best results in red and blue.}
\label{table:DN173}
\begin{center}
\resizebox{\linewidth}{!}{
\begin{small}
\begin{sc}
\begin{tabular}{l|ccccccc}
\toprule
& P2R & P2S & R2P & R2S & S2P & S2R & Avg. \\
\midrule
SIMPLE    &59.6	&53.4&	53.5&	56.3&	52.5&	59.6&	55.9 \\
GCD    & 45.9&	33.7	&41.1&	38.6&	39.0	&48.5	&41.3 \\
ORCA    & 36.7&	30.4	&32.1&	30.9&	27.7&	31.1&	31.5 \\
DCC    & 53.6&	42.7&	45.9&	45.6&	43.7&	50.9	&47.5\\
DANCE    & 60.2&	51.1&	\cellcolor{red!20}54.1	&52.5&	\cellcolor{blue!20}53.9	&61.8	&55.7  \\
OVAnet    & 61.3&	50.6&	\cellcolor{blue!20}53.9&	\cellcolor{blue!20}52.6&	52.8&	61.3&	55.6 \\
UniOT    & 54.3	&45.4&	52.3&	48.1	&50.5	&55.1&	51.2  \\
NCDDA&61.2&	49.1&	45.2&	50.4&	48.8&	57.5&	52.6\\
SAN&\cellcolor{blue!20}62.6&	\cellcolor{blue!20}53.4	&49.1&	51.0&	47.8&	\cellcolor{blue!20}64.2&	54.8\\
GLC    & 62.1&	51.4	&53.3&	\cellcolor{red!20}54.4	&51.9	&60.8	&\cellcolor{blue!20}56.1  \\
\method    & \cellcolor{red!20}67.9	&\cellcolor{red!20}54.2	&53.7	&51.7	&\cellcolor{red!20}54.8&	\cellcolor{red!20}70.4&	\cellcolor{red!20}59.0\\
\bottomrule
\end{tabular}
\end{sc}
\end{small}
}
\end{center}
\end{table}

\begin{table}[h]
\caption{H-score (\%) of dataset DomainNet (105/240) on different pairs of domain. We color the best and second-best results in red and blue.}
\label{table:DN105}
\begin{center}
\resizebox{\linewidth}{!}{
\begin{small}
\begin{sc}
\begin{tabular}{l|ccccccc}
\toprule
& P2R & P2S & R2P & R2S & S2P & S2R & Avg. \\
\midrule
SIMPLE    & 65.0	&55.1	&57.0	&56.9&	52.8	&59.9&	57.8 \\
GCD    & 48.0	&30.6&	40.3	&37.2&	38.3&	52.5&	41.3 \\
ORCA    & 33.2	&37.1	&29.4&	29.2	&30.4&	42.2	&33.7 \\
DCC    & 57.0&	41.4&	43.8&	42.4&	44.8&	55.6&	47.7 \\
DANCE    & 64.7	&54.5	&\cellcolor{red!20}57.2	&\cellcolor{blue!20}54.4&	\cellcolor{blue!20}56.2&	65.4	&\cellcolor{blue!20}58.8  \\
OVAnet    & 64.7	&\cellcolor{blue!20}56.0&	55.4	&53.8	&54.5&	\cellcolor{blue!20}66.1	&58.5 \\
UniOT    & 54.5	&50.4	&48.1	&48.6&	47.6&	55.6&	50.8 \\
NCDDA&\cellcolor{blue!20}67.0&	54.3&	50.7&	47.2&	49.2&	64.1&	55.7\\
SAN&63.7&	53.9&	49.9&	49.9&	51.3&	60.9	&55.0\\
GLC    &62.8&	50.7&	53.4	&50.8	&50.8&	62.3&	55.7 \\
\method    & \cellcolor{red!20}71.3	&\cellcolor{red!20}56.5	&\cellcolor{blue!20}56.6	&\cellcolor{red!20}54.9	&\cellcolor{red!20}58.2&	\cellcolor{red!20}70.8	&\cellcolor{red!20}61.5\\
\bottomrule
\end{tabular}
\end{sc}
\end{small}
}
\end{center}
\end{table}

\begin{table*}[h]
\caption{H-score (\%) of dataset OfficeHome (45/20) on different pairs of domain. We color the best and second-best results in red and blue.}
\label{table:oh45}
\begin{center}
\resizebox{\textwidth}{!}{
\begin{small}
\begin{sc}
\begin{tabular}{l|ccccccccccccc}
\toprule
& A2C & A2P & A2R & C2A & C2P & C2R & P2A & P2C & P2R & R2A & R2C & R2P &Avg.\\
\midrule
SIMPLE    &55.4&	68.2&	74.9&	58.7	&69.4	&63.1&	47.8&	47.1&	69.6&	59.6	&55.6&	72.9&	62.3 \\
GCD    & 39.4	&56.1&	64.4&	37.0&	45.8	&47.3&	39.4	&32.2	&60.8&	34.2&	43.2&	62.0&	48.7 \\
ORCA    & 44.7	&60.8&	50.9&	43.6	&53.9	&44.4&	44.4&	42.5	&61.4&	43.1&	43.2&	45.5	&48.9 \\
DCC    & 60.2	&69.3	&68.2	&46.3&	74.7&	69.9&	48.5&	\cellcolor{blue!20}61.9&	69.3	&52.6&	\cellcolor{blue!20}64.2&	76.0&	63.6 \\
DANCE    & 56.6	&72.5&	78.0	&\cellcolor{red!20}65.7&	70.4&	68.5&	\cellcolor{red!20}59.5	&55.8&	70.2&	65.1	&57.7&	65.4&	65.7  \\
OVAnet    & 57.2	&71.1	&\cellcolor{blue!20}78.2	&\cellcolor{blue!20}63.5&	67.6	&\cellcolor{blue!20}73.7	&\cellcolor{blue!20}55.0&	48.4&	\cellcolor{blue!20}73.7&	\cellcolor{blue!20}66.9&	59.6&	77.2	&\cellcolor{blue!20}66.4  \\
UniOT    &58.1&	70.4&	72.9	&57.3	&69.3&	65.6	&54.3	&55.1&	59.3&	\cellcolor{red!20}67.5	&63.1&	76.9&	64.4 \\
NCDDA&\cellcolor{red!20}64.4&	\cellcolor{blue!20}76.3&	64.5&	53.5	&\cellcolor{blue!20}73.4	&55.4&	48.7&	54.6&	60.1	&60.2	&59.0&	\cellcolor{blue!20}81.5&	63.2\\
SAN&57.8&	69.8&	71.8&	53.9&	72.1&	71.0&	52.0&	55.9&	63.3	&61.7&	61.9	&76.9&	64.3\\
GLC    &59.3&	71.9&	61.9&	57.7&	73.3&	61.7&	54.3&	58.7&	66.6&	64.0&	56.7&	73.5&	65.2\\
\method    & \cellcolor{blue!20}63.1	&\cellcolor{red!20}82.5&	\cellcolor{red!20}79.5&	48.3&	\cellcolor{red!20}83.1&	\cellcolor{red!20}75.6&	51.8&	\cellcolor{red!20}64.7&	\cellcolor{red!20}75.2&	54.8&	\cellcolor{red!20}67.8&	\cellcolor{red!20}84.5&	\cellcolor{red!20}69.4\\
\bottomrule
\end{tabular}
\end{sc}
\end{small}
}
\end{center}
\end{table*}

\begin{table*}[h]
\caption{H-score (\%) of dataset OfficeHome (33/32) on different pairs of domain. We color the best and second-best results in red and blue.}
\label{table:oh33}
\begin{center}
\resizebox{\textwidth}{!}{
\begin{small}
\begin{sc}
\begin{tabular}{l|ccccccccccccc}
\toprule
& A2C & A2P & A2R & C2A & C2P & C2R & P2A & P2C & P2R & R2A & R2C & R2P &Avg.\\
\midrule
SIMPLE    & 59.9	&72.2	&77.8&	64.3	&73.5	&65.1	&57.0&	55.8&	70.1	&63.4&	58.3&	72.9	&66.0 \\
GCD    & 44.2	&46.7	&59.6&	40.8&	46.6&	50.1&	30.7	&28.2&	55.1&	39.3&	45.5&	62.0	&47.5 \\
ORCA    & 47.4	&62.5	&44.9	&44.0&	52.3	&46.7&	38.4&	29.3&	52.5&	52.2&	48.5&	63.3&	48.9 \\
DCC    & 64.1	&74.9	&71.1&	48.5&	\cellcolor{blue!20}78.1&	70.6&	51.1	&58.6&	68.0	&48.8&	60.5&	\cellcolor{blue!20}82.9	&65.0 \\
DANCE    & 52.9	&70.8&	74.3	&\cellcolor{blue!20}66.9&	69.3&	73.4&	\cellcolor{red!20}59.5	&54.2&	70.9	&\cellcolor{blue!20}67.0	&52.8&	72.0	&65.6  \\
OVAnet    & 57.9	&72.5&	\cellcolor{blue!20}78.4&	\cellcolor{red!20}67.6&	69.6&	73.8	&\cellcolor{blue!20}57.8&	53.2	&\cellcolor{blue!20}74.9	&\cellcolor{red!20}69.4&	\cellcolor{blue!20}61.1&	82.2&	\cellcolor{blue!20}68.6  \\
UniOT    &59.5&	72.1	&71.0&	57.1	&74.2&	63.2&	52.4	&56.8&	67.5&	65.8&	58.9&	76.9&	64.9 \\
NCDDA&\cellcolor{blue!20}64.9	&\cellcolor{blue!20}76.8&	64.8&	56.7	&\cellcolor{blue!20}78.1	&55.7&	49.7&	55.2	&60.1	&61.2	&59.8&	82.4&	64.3\\
SAN&60.5	&70.8	&72.2&	56.3&	71.6	&69.4&	54.7&	59.0	&63.3&	61.5&	59.5	&77.4	&65.0\\
GLC    & 64.4&	71.8	&\cellcolor{red!20}78.9&	51.3&	76.0&	\cellcolor{red!20}76.6&	48.2&	\cellcolor{red!20}63.5&	\cellcolor{red!20}80.4	&55.9	&60.0	&75.4&	68.2\\
\method    & \cellcolor{red!20}66.5	&\cellcolor{red!20}83.4&	77.7&	50.3&	\cellcolor{red!20}82.2&	\cellcolor{blue!20}75.4&	53.2	&\cellcolor{blue!20}62.6&	74.0	&56.2&	\cellcolor{red!20}66.2&	\cellcolor{red!20}84.2	&\cellcolor{red!20}69.6\\
\bottomrule
\end{tabular}
\end{sc}
\end{small}
}
\end{center}
\end{table*}

\begin{table*}[h]
\caption{H-score (\%) of dataset OfficeHome (20/45) on different pairs of domain. We color the best and second-best results in red and blue.}
\label{table:oh20}
\begin{center}
\resizebox{\textwidth}{!}{
\begin{small}
\begin{sc}
\begin{tabular}{l|ccccccccccccc}
\toprule
& A2C & A2P & A2R & C2A & C2P & C2R & P2A & P2C & P2R & R2A & R2C & R2P &Avg.\\
\midrule
SIMPLE    & 58.7&	70.5	&76.6	&63.8&	72.9	&64.6	&57.0	&56.5	&69.1	&63.4	&57.7	&72.9&	65.4\\
GCD    & 37.8	&51.8	&59.3	&46.6	&46.3	&49.4&	26.0	&18.3	&54.3	&49.9	&45.2	&69.4	&48.1 \\
ORCA    & 42.4	&55.4	&57.0&	60.2&	36.7	&62.2	&40.7	&44.1&	55.9&	47.3&	22.2&	58.0	&49.9 \\
DCC    & 63.9&	74.4&	71.0&	47.6&	\cellcolor{blue!20}78.4&	70.1	&50.9	&58.5&	67.7&	50.2&	59.8&	\cellcolor{blue!20}81.3&	64.7\\
DANCE    & 55.7	&72.0&76.4	&\cellcolor{blue!20}66.7&	69.5&	71.4	&\cellcolor{red!20}60.7	&51.4&	73.5	&\cellcolor{red!20}71.3&	56.9	&72.9&	67.1  \\
OVAnet    & 58.7&	72.2&	76.7&	\cellcolor{red!20}66.9&	70.3&	73.5&	\cellcolor{blue!20}58.4&	56.3&	\cellcolor{red!20}77.5&	\cellcolor{blue!20}69.5&	\cellcolor{blue!20}62.4&	74.6&	68.7  \\
UniOT    & 58.0&	72.2&	68.1&	57.1&	75.6&	63.6&	49.4	&\cellcolor{blue!20}59.2&	67.5&	66.0	&59.1&	77.7	&64.8 \\
NCDDA&65.9	&78.3&	66.0&	58.4	&76.6&	57.3	&51.5&	59.1&	60.1	&61.2	&61.9	&78.9&	65.0\\
SAN&63.4&	78.3	&74.1&	56.9	&76.0&	73.0	&57.8	&56.5	&67.9	&61.8&	59.4&	78.9	&67.2\\
GLC    & \cellcolor{red!20}69.2&	\cellcolor{red!20}85.8&	\cellcolor{red!20}83.7&	64.0	&75.2	&\cellcolor{red!20}77.8	&55.2&	52.9&	74.5&	62.3&	57.2&	72.2&	\cellcolor{blue!20}69.3\\
\method    & \cellcolor{red!20}69.2	&\cellcolor{blue!20}79.0	&\cellcolor{blue!20}78.0&	52.6&	\cellcolor{red!20}81.7	&\cellcolor{blue!20}73.7&	53.1	&\cellcolor{red!20}68.1&	\cellcolor{blue!20}76.8&	56.6	&\cellcolor{red!20}69.6&	\cellcolor{red!20}82.6&	\cellcolor{red!20}70.2\\
\bottomrule
\end{tabular}
\end{sc}
\end{small}
}
\end{center}
\end{table*}

\clearpage

\begin{table*}[h]
\caption{Average seen accuracy (\%), unseen accuracy (\%), and H-score (\%) of 70\% seen/unseen splits on dataset Office, OfficeHome, VisDA, and DomainNet. We color the best and second-best results in red and blue.}
\label{table:main2-70}
\begin{center}
\resizebox{\textwidth}{!}{
\begin{small}
\begin{sc}
\begin{tabular}{l|ccc|ccc|ccc|ccc}
\toprule
& \multicolumn{3}{c|}{Office (21/10)} & \multicolumn{3}{c|}{OfficeHome (45/20)} & \multicolumn{3}{c|}{VisDA (8/4)} & \multicolumn{3}{c}{DomainNet (240/105)} \\
& seen & unseen & H-score & seen & unseen & H-score & seen & unseen & H-score & seen & unseen & H-score \\
\midrule
SIMPLE    & 64.6&65.3&64.9 & 65.3&59.5&62.3 & 57.3&53.7&55.4 & 69.1&43.2&53.2 \\
GCD    & 67.6 & 58.4 & 62.6 & 50.9 & 46.7 & 48.7 & 31.8 & 30.6 & 31.2 & 39.9 & 31.2 & 35.0  \\
ORCA    & 74.0 & 47.6 & 57.9 & 69.8 & 37.6 & 48.9 & 65.2 & 20.6 & 31.3 & 58.1 & 19.3 & 28.9\\
DCC    & 74.6 & 71.1 & 72.8 & 66.1 & 61.3 & 63.6 & \cellcolor{blue!20}73.9 & 51.5 & 60.7 & 45.7 & 45.4 & 45.5 \\
DANCE    & 79.2 & \cellcolor{blue!20}71.9 & 75.4 & \cellcolor{red!20}74.2 & 58.9 & 65.7 & 64.9 & 51.2 & 57.2 & \cellcolor{blue!20}67.9 & 48.1 & \cellcolor{blue!20}56.3 \\
OVAnet    & 78.7 & 68.5 & 73.2 & 67.9 & \cellcolor{blue!20}65.1 & \cellcolor{blue!20}66.4 & 56.5 & 63.7 & 59.9 & 60.2 & 49.2 & 54.2 \\
UniOT    & 81.7 & 71.2 & 76.1 & 70.6 & 59.3 & 64.4 & 72.3 & 54.2 & \cellcolor{blue!20}62.0 & 49.1 & 42.7 & 45.7 \\
NCDDA&\cellcolor{blue!20}91.6&71.5&80.3&66.9&59.8&63.2&67.3&49.8&57.2&58.1&44.1&50.1\\
SAN&\cellcolor{red!20}93.1&71.0&\cellcolor{blue!20}80.5&68.8&60.4&64.3&69.3&54.8&61.2&67.3&44.0&53.2\\
GLC    & 89.4 & 65.4 & 75.5 & \cellcolor{blue!20}73.8 & 58.4 & 65.2 & 58.6 & \cellcolor{blue!20}64.1 & 61.2 & 61.9 & \cellcolor{red!20}49.3 & 54.9 \\
\method    & 90.9 & \cellcolor{red!20}79.2 & \cellcolor{red!20}84.7 & 68.6 & \cellcolor{red!20}70.3 & \cellcolor{red!20}69.4 & \cellcolor{red!20}76.8 & \cellcolor{red!20}65.1 & \cellcolor{red!20}70.5 & \cellcolor{red!20}69.8 & \cellcolor{red!20}49.3 & \cellcolor{red!20}57.8 \\
\bottomrule
\end{tabular}
\end{sc}
\end{small}
}
\end{center}
\end{table*}

\begin{table*}[h]
\caption{Average seen accuracy (\%), unseen accuracy (\%), and H-score (\%) of 30\% seen/unseen splits on dataset Office, OfficeHome, VisDA, and DomainNet. We color the best and second-best results in red and blue.}
\label{table:main2-30}
\begin{center}
\resizebox{\textwidth}{!}{
\begin{small}
\begin{sc}
\begin{tabular}{l|ccc|ccc|ccc|ccc}
\toprule
& \multicolumn{3}{c|}{Office (10/21)} & \multicolumn{3}{c|}{OfficeHome (20/45)} & \multicolumn{3}{c|}{VisDA (4/8)} & \multicolumn{3}{c}{DomainNet (105/240)} \\
& seen & unseen & H-score & seen & unseen & H-score & seen & unseen & H-score & seen & unseen & H-score \\
\midrule
SIMPLE    & 86.3&71.8&78.3 & 72.0&59.9&65.4 & 58.4&45.1&50.9 & 76.6&46.4&57.8 \\
GCD    & 49.7 & 69.6 & 58.0 & 42.1 & 56.1 & 48.1 & 29.6 & 21.9 & 25.2 & 40.4 & 42.3 & 41.3  \\
ORCA    & 71.2 & 55.4 & 62.3 & 67.1 & 39.7 & 49.9 & 69.3 & 22.4 & 33.9 & 68.7 & 22.3 & 33.7 \\
DCC    & 73.8 & 76.6 & 75.2 & 72.1 & 58.9 & 64.7 & 65.8 & 49.8 & 56.7 & 57.5 & 40.7 & 47.7\\
DANCE    & 83.5 & 60.1 & 69.9 & 69.1 & \cellcolor{blue!20}65.3 & 67.1 & 75.2 & 35.7 & 48.4 & \cellcolor{blue!20}71.3 & \cellcolor{blue!20}50.1 & \cellcolor{blue!20}58.8 \\
OVAnet    & 74.7 & 75.9 & 75.3 & 72.7 & 65.2 & 68.7 & 62.3 & \cellcolor{blue!20}58.2 & 60.2 & 70.3 & \cellcolor{blue!20}50.1 & 58.5 \\
UniOT    & 90.3 & \cellcolor{blue!20}77.5 & \cellcolor{blue!20}83.4 & 73.9 & 57.6 & 64.8 & \cellcolor{blue!20}75.7 & 49.4 & 59.8 & 61.6 & 43.3 & 50.8\\
NCDDA&\cellcolor{blue!20}93.4&72.5&81.7&71.8&59.4&65.0&70.8&51.0&59.3&70.1&46.2&55.7\\
SAN&\cellcolor{red!20}95.7&71.8&82.0&\cellcolor{blue!20}77.1&59.5&67.2&74.3&52.4&61.5&68.7&45.8&55.0\\
GLC    & 91.7 & 66.9 & 77.3 &\cellcolor{red!20} 77.7 & 62.6 & \cellcolor{blue!20}69.3 & \cellcolor{red!20}76.3 & 53.2 & \cellcolor{blue!20}62.7 & 65.7 & 48.3 & 55.7 \\
\method   & 88.8 & \cellcolor{red!20}82.6 & \cellcolor{red!20}85.6 & 70.4 & \cellcolor{red!20}70.0 & \cellcolor{red!20}70.2 & 68.9 & \cellcolor{red!20}73.4 & \cellcolor{red!20}71.1 & \cellcolor{red!20}72.5 & \cellcolor{red!20}53.5 & \cellcolor{red!20}61.5\\
\bottomrule
\end{tabular}
\end{sc}
\end{small}
}
\end{center}
\end{table*}

\end{document}